\documentclass[lettersize,journal]{IEEEtran}

\usepackage[cmex10]{amsmath} 
\usepackage{amsfonts,stmaryrd,acronym,amssymb,amsthm,dsfont}
\usepackage{algorithm}
\usepackage{algorithmic}
\usepackage{graphicx,paralist,mdframed}
\usepackage{graphics}
\usepackage{bm}
\usepackage{url}
\usepackage{cite}
\usepackage{caption}
\usepackage{subcaption}
\usepackage{hyperref}
\usepackage{flushend}
\usepackage{enumerate}
\usepackage{pdflscape}
\usepackage[dvipsnames]{xcolor}
\usepackage{mathrsfs}
\usepackage{lipsum}
\usepackage{diagbox}
\usepackage{soul}
\usepackage{kantlipsum}
\usepackage{ragged2e} 
\usepackage{booktabs}

\usepackage[bbgreekl]{mathbbol}
\DeclareSymbolFontAlphabet{\mathbbm}{bbold}
\DeclareSymbolFontAlphabet{\mathbb}{AMSb}%

\graphicspath{{Graphics/}}
\definecolor{ColorHighlight}{rgb}{1,0,0}

\makeatletter
\def\blfootnote{\xdef\@thefnmark{}\@footnotetext}
\makeatother
\newcommand{\indi}[1]{\ensuremath{\mathds{1}}}

\newcounter{mytempeqcounter}

\theoremstyle{plain}

\theoremstyle{definition}

\theoremstyle{remark}
\newtheorem{remark}{Remark}

\allowdisplaybreaks
\allowbreak

\makeatletter%
\if@twocolumn%
\else
\fi%
\makeatother%

\newcommand{\indic}[1]{\ensuremath{\mathds{1}}}

\renewcommand{\leq}{\leqslant} 
\renewcommand{\geq}{\geqslant} 

\acrodef{ACDIS}[ACDIS]{Adaptive Communication Decision and Information Systems}
\acrodef{AEP}{Asymptotic Equipartition Property}
\acrodef{AWGN}{additive white gaussian noise}
\acrodef{AVC}[AVC]{Arbitrarily Varying Channel}
\acrodefplural{AVC}{Arbitrarily Varying Channels}
\acrodef{PIR-PNSI}{Private Information Retrieval with Private Noisy Side Information}
\acrodef{BER}{bit-error-rate}
\acrodef{BEC}{Binary Erasure Channel}
\acrodefplural{BEC}{Binary Erasure Channels}
\acrodef{BSC}{Binary Symmetric Channel}
\acrodefplural{BSC}{Binary Symmetric Channels}
\acrodef{BPSK}{Binary Phase-Shift Keying}
\acrodef{BICM}[BICM]{Bit-Interleaved Coded-Modulation}
\acrodef{CDF}[CDF]{Cumulative Distribution Function}
\acrodef{CGF}[CGF]{Cumulant Generating Function}
\acrodef{CLT}[CLT]{Central Limit Theorem}
\acrodef{DMC}[DMC]{Discrete Memoryless Channel}
\acrodefplural{DMC}{Discrete Memoryless Channels}
\acrodef{DMS}[DMS]{Discrete Memoryless Source}
\acrodef{ERM}[ERM]{Empirical Risk Minimization}
\acrodef{FER}[FER]{Frame Error Rate}
\acrodef{ICA}[ICA]{Independent Component Analysis}
\acrodef{iid}[i.i.d.]{independent and identically distributed}
\acrodef{IoT}[IoT]{Internet of Things}
\acrodef{KKT}[KKT]{Karush-Kuhn Tucker}
\acrodef{LASSO}[LASSO]{Least Absolute Shrinkage and Selection Operator}
\acrodef{LPD}[LPD]{Low Probability of Detection}
\acrodef{LDPC}[LDPC]{Low-Density Parity-Check}
\acrodef{CSCG}[CSCG]{Circularly Symmetric Complex Gaussian Distribution}
\acrodef{LLMS}[LLMS]{Linear Least Mean Square}
\acrodef{LMS}[LMS]{Least Mean Square}
\acrodef{MAC}[MAC]{Multiple-Access Channel}
\acrodef{ADSI}[ADSI]{Action-Dependent State Information}
\acrodef{MGF}[MGF]{Moment Generating Function}
\acrodef{MLC}[MLC]{multi-Level Coding}
\acrodef{MLE}[MLE]{maximum Likelihood Estimate}
\acrodef{MIMO}[MIMO]{multiple-Input Multiple-Output}
\acrodef{MISO}{multiple-Input Single-Output}
\acrodef{MSD}[MSD]{Multi-Stage Decoding}
\acrodef{MMSE}[MMSE]{minimum Mean-Square Error}
\acrodef{PAC}[PAC]{Probably Approximately Correct}
\acrodef{PCA}[PCA]{Principal Component Analysis}
\acrodef{PDF}[PDF]{Probability Density Function}
\acrodefplural{PDF}{Probability Density Functions}
\acrodef{PMF}[PMF]{Probability Mass Function}
\acrodefplural{PMF}{Probability Mass Functions}
\acrodef{PPM}[PPM]{Pulse Position Modulation}
\acrodef{PSD}{Power Spectral Density}
\acrodef{PSK}{Phase Shift Keying}
\acrodef{QKD}{Quantum Key Distribution}
\acrodef{ROC}{Receiver Operating Characteristic}
\acrodef{CVQKD}{Continuous-Variable \ac{QKD}}
\acrodef{QPSK}{Quadrature Phase-Shift Keying}
\acrodef{RV}{random variable}
\acrodef{SIMO}{Single-Input Multiple-Output}
\acrodef{SNR}{signal-to-noise ratio}
\acrodef{SVM}[SVM]{Support Vector Machine}
\acrodef{TPCP}{Trace-Preserving Completely-Positive}
\acrodef{wrt}[w.r.t.]{with respect to}
\acrodef{WSS}{Wide Sense Stationary}
\acrodef{RHS}{Right Hand Side}
\acrodef{LHS}{Left Hand Side}
\acrodef{PIR}{Private Information Retrieval}
\acrodef{MDS}{Maximum Distance Separable}
\acrodef{LLN}{law of Large Numbers}
\acrodef{DFRC}{dual-Function Radar Communication}
\acrodef{ISAC}{integrated sensing and communication}
\acrodef{RadCom}{Joint Radar and Communicatins}
\acrodef{PLS}[PLS]{Physical Layer Security}
\acrodef{RL}{reinforcement learning}
\acrodef{POCS}{projections onto convex sets}
\acrodef{SINR}{signal-to-interference-ratio}
\acrodef{RNN}{recurrent neural network}
\acrodef{BS}{base station}
\acrodef{MISO}{multiple-input-single-output}
\acrodef{MIMO}{multiple-input-multiple-output}
\acrodef{mmWave}{millimeter wave}
\acrodef{RF}{Radio frequency}
\acrodef{PS}{Phase shifter}
\acrodef{TTD}{true time delayer}
\acrodef{FDD}{frequency division duplex}
\acrodef{TDD}{time division duplex}
\acrodef{NN}{neural network}
\acrodef{CSI}{channel state information}
\acrodef{GAN}{generative Adversarial Network}
\acrodef{ULA}{uniform linear array}
\acrodef{BiCNN}{bi-directional convolutional NN}
\acrodef{EDN}{encoder-decoder network}
\acrodef{ISI}{inter-user interference}
\acrodef{KD-EDN}{knowledge-distillation-based EDN}
\acrodef{DNN}{deep neural network}
\acrodef{KD}{knowledge distillation}
\acrodef{AOA}{angle of arrival}
\acrodef{AOD}{angle of departure}
\acrodef{URLLC}{ultra-reliable low-latency communication}
\acrodef{L2O}{learning-to-optimize}
\acrodef{MRT}{maximum ratio transmission}
\acrodef{BRB}{branch-reduce-bound}
\acrodef{DL}{deep learning}
\acrodef{CNN}{convolutional neural network}
\acrodef{WMMSE}{weighted MMSE}
\acrodef{GNN}{graphical neural network}
\acrodef{IRS}{intelligent reflect surface}
\acrodef{CS}{compressed sensing}
\acrodef{SVD}{singular value decomposition}
\acrodef{ADMM}{alternating direction method of multipliers}
\acrodef{QoS}{quality of Service}
\acrodef{KDL}{KKT-guided dual learning}
\acrodef{PASS}{pinching antenna systems}
\acrodef{LSTM}{long short term memory}
\acrodef{SCA}{successive convex optimization}
\acrodef{RIS}{reconfigurable intelligent surface}
\acrodef{CL}{curriculum learning}
\acrodef{PE}{permutation equivariance}
\acrodef{MHSA}{multi-head self-attention}
\acrodef{MLP}{multi-layer perceptron}
\acrodef{PGA}{projected gradient ascent}
\acrodef{PGD}{projected gradient descent}
\acrodef{TBPTT}{truncated backpropagation through time}
\acrodef{SOTA}{state-of-the-art}
\acrodef{IUI}{inter-user interference}
\acrodef{PC}{positional coding}
\acrodef{FC}{fully-connected}
\acrodef{TN}{token-wise normalization}
\acrodef{HPE}{hierarchical permutation equivariance}
\acrodef{SGD}{stochastic gradient descent}
\acrodef{JRCB}{joint radar-communication beamforming}
\acrodef{RCS}{radar cross section}
\acrodef{LS}{least square}
\acrodef{SDP}{semidefinite program}
\acrodef{SAO}{semi-amortized learning-to-optimize}
\acrodef{SA-L2O}{semi-amortized learning-to-optimize}
\acrodef{MT-SAO}{masked-Transformer-based SAO}
\acrodef{MT-SAO-lift}{masked-Transformer-based SAO-lift}
\acrodef{DT-SAO-lift}{dedicated-Transformer-based SAO-lift}
\acrodef{SAO-ISAC}{SAO-for-ISAC}
\acrodef{LLM}{large language model}
\acrodef{SALLO}{semi-amortized lifted learning-to-optimize}
\acrodef{SALO}{semi-amortized learning-to-optimize}
\acrodef{SALLO-M}{semi-amortized lifted learning-to-optimize with masking}

\begin{document}



\title{A Semi-amortized Lifted Learning-to-Optimize Masked (SALLO-M) Transformer Model for Scalable and Generalizable Beamforming}

\author{
\IEEEauthorblockN{Yubo Zhang, Xiao-Yang Liu, and Xiaodong Wang}\\
\thanks{Y.~Zhang and X.~Liu and X.~Wang are with the Department of Electrical Engineering, Columbia University, New York, NY 10027.}
}
\maketitle
\date{}

\begin{abstract}
\label{sec:Abstract}
We develop an unsupervised deep learning framework for real-time scalable and generalizable downlink beamforming in multi-user multiple-input single-output (MU-MISO) systems. The proposed semi-amortized lifted learning-to-optimize (SALLO) framework employs a multi-layer Transformer to iteratively refine an auxiliary variable and the beamformer solution, with a few projected gradient ascent steps at each layer. A key feature of our SALLO Transformer model is that it can handle varying numbers of users and antennas, enabled by a user-antenna dual tokenization and a structured sample/attention masking scheme, leading to generalization across different configurations without retraining. To improve convergence and robustness, we introduce three training strategies: (a) sliding-window training to stabilize gradient propagation, (b) curriculum learning with random masking to enable user-antenna configuration generalization and prevent poor early-stage convergence, and (c) sample replay to mitigate catastrophic forgetting during multi-stage training. Ablation studies validate several key architecture designs and show that the enhanced training scheme improves both generalizability and solution quality. Simulation results over both Gaussian and sparse channels show that the proposed scheme consistently outperforms existing deep learning baselines across diverse system configurations and channel conditions. The performance gain becomes more pronounced in overloaded regimes, highlighting improved robustness under challenging scenarios. Furthermore, our scheme surpasses the WMMSE benchmark in underloaded systems and even in overloaded systems when the overloading factor is below certain threshold. These gains are achieved with fast inference and a substantially more lightweight model than wireless foundation models.



\end{abstract}


\section{Introduction} \label{sec_intro}

Next-generation wireless systems are expected to operate at higher carrier frequencies with larger antenna arrays, calling for scalable architectures and low-latency signal processing. Among various physical-layer techniques, real-time downlink beamforming --- where the \ac{BS} continuously updates its transmit beamformers according to time-varying channels --- has become a key enabler for high spectral efficiency and has attracted significant research attention. State-of-the-art iterative methods, such as the \ac{BRB} algorithm~\cite{bjornson2013optimal} and the \ac{WMMSE} method~\cite{shi2011iteratively}, can achieve near-optimal performance, but are often computationally prohibitive in large-scale systems. In contrast, low-complexity schemes such as the \ac{MRT}~\cite{lozano2006optimum} and the linear MMSE (LMMSE) beamforming \cite{bjornson2014optimal} offer fast solutions at the cost of substantial performance degradation. Recently, \ac{DL} approaches, especially Transformer-based models, have been investigated to enable high-quality beamforming with improved scalability and real-time adaptability.

\subsection{Deep Learning for Downlink Beamforming} \label{intro_dl_opt}

Deep learning for beamforming and precoding in multi-user systems has been under rapid development in recent years. Early works relied mainly on lightweight architectures such as fully-connected neural networks (FCNNs) \cite{sohrabi2021deep,zhang2025encoderdecodernetworkbeamformingsparse}, CNNs~\cite{elbir2019hybrid}, and RNN variants~\cite{liu2022learning} to learn compact CSI representations and/or directly predict beamformers with reduced feedback overhead. Although computationally efficient, these models typically require retraining when the system configuration changes and scale poorly to larger systems due to the limited expressive power of the models. On the other hand, a GNN-based model is typically size-generalizable~\cite{kim2022bipartite,li2024gnn}, but its expressive power is often limited by the lack of global dependency modeling. To tackle this issue, Transformer-based models are introduced to capture richer global structural information in precoding-related tasks \cite{zhang2024transformer,ting2024adaptive}. However, a vanilla Transformer tends to treat channel modeling as a generic one-dimensional token-mixing task and only adapts to different user numbers, which leads to non-generalizability to other varying parameters (e.g., antenna numbers, pilot length) in dynamic channels~\cite{li2024hpe,duan2025learning}. To tackle this problem, \cite{duan2025learning} proposes a Transformer model with GNN-guided structural constraints to enable global size-generalizability, at the cost of reduced expressive power. More recently, several works have explored fine-tuning pretrained foundation models~\cite{xu2025llm,li2025bert4beam}, or training wireless foundation models from scratch~\cite{wen2026wifo}, to further improve the multi-scenario and multi-task adaptability for physical-layer optimizations. However, these schemes typically incur a much higher training cost and model size, making them less attractive for resource-constrained deployments.


\subsection{Learning Techniques for Optimization} \label{learning_opt_paradigm}

Several learning techniques have been proposed to solve non-convex optimization problems. To begin with, the \textbf{\ac{L2O}/amortized optimization} paradigm shifts the computational burden from online iterative optimization to offline training. Instead of solving each problem instance from scratch, L2O learns an optimizer from the data, exploiting recurring task structures for fast inference. Early works formulated optimizer design itself as a learning problem, for example by using reinforcement learning to discover optimization policies~\cite{li2017learning}. Subsequent studies improved scalability and transferability by developing learned optimizers that generalize to larger and unseen tasks~\cite{wichrowska2017learned}. More recently, L2O has been applied to physical-layer optimization problems. For example,~\cite{lavi2023learn} proposes an interpretable L2O-based hybrid precoding framework with reliable convergence and improved robustness under noisy CSI, while~\cite{johnston2023curriculum} develops a bidirectional CNN architecture for beamforming optimization. Meanwhile, the semi-amortized paradigm enhances amortized optimization by combining a learned initializer with a small number of refinement steps, preserving fast inference while improving instance-wise optimality~\cite{kim2018semi}.


Another learning technique is the \textbf{masking} operation, which has been widely used as both a training strategy and a structural constraint to make learning easier, more stable, and more efficient. In self-supervised pretraining, masked reconstruction provides a strong surrogate objective. For example, masked autoencoders (MAE) reconstruct masked image patches to improve representation learning and training efficiency~\cite{he2022masked}, while masked language modeling (MLM) in BERT predicts hidden tokens from bidirectional context and has become a standard pretraining strategy~\cite{devlin2019bert}. In sequence generation, causal masking prevents attention to future tokens and forms the basis of auto-regressive Transformer decoding~\cite{vaswani2017attention}. Masking can also shape optimization by restricting attention patterns or search spaces. For example, \cite{fan2021mask} introduces dynamic learnable masking to better capture locality, and Mask2Former~\cite{cheng2022masked} uses masked attention to constrain cross-attention within predicted mask regions, thus improving local feature extraction in vision segmentation.

\subsection{Contributions and Outline} \label{contri_outline}


Existing learning-based beamforming methods face the fundamental difficulty of simultaneously achieving scalability, broad generalizability, and a lightweight model with fast inference. In addition, their performance tends to significantly deteriorate in more challenging scenarios, such as user-overloaded systems, indicating insufficient robustness. To address these issues, we propose a \textbf{semi-amortized lifted learning-to-optimize masked (SALLO-M) Transformer model}, which provides strong scalability, improved generalizability within a size upper bound, and robust sum-rate performance across varying system settings, including several challenging scenarios. The main contributions are summarized as follows.
\begin{itemize}
    \item \textbf{Semi-amortized lifted learning-to-optimize (SALLO) framework:} We propose a multi-layer optimizer architecture in which an auxiliary variable and the beamformer are jointly updated through residual connections at each Transformer layer. This lifted representation expands the effective feature space and enhances the model expressivity. To further improve solution quality, each layer's output is refined by a small number of projected gradient ascent steps, following the semi-amortized learning paradigm that combines fast amortized inference with instance-wise optimization.

    \item \textbf{Dual tokenization and masked Transformer architecture:} We adopt a user-antenna dual tokenization scheme and design a masked Transformer architecture for downlink systems with varying and bounded numbers of users and antennas. By introducing structured sample/attention masking based on the dual tokenization, the resulting model with fixed input-output dimensions can be efficiently trained on heterogeneous channel configurations and generalize across varying settings without retraining, enabling real-time adaptation to dynamic environments.

    \item \textbf{Enhanced training strategies:} We develop three complementary training strategies to improve convergence and generalizability of SALLO-M Transformer model: (i)~a sliding-window training method that stabilizes gradient flow and reduces memory consumption when the number of Transformer layers is large; (ii)~a curriculum learning (CL) scheme with random masking that progressively exposes the model to samples with increasingly challenging configurations, which both enables generalizability and prevents a poor early-stage convergence; and (iii)~a sample replay mechanism that mitigates catastrophic forgetting in the curriculum stages.

    \item \textbf{Performance evaluations:} Comprehensive ablation studies validate the effectiveness of key architectural designs in the proposed model, and confirm that the enhanced CL-based training scheme improves both generalizability and solution quality. The results further demonstrate that across a wide range of system configurations and channel conditions, the proposed scheme consistently outperforms the existing deep learning baselines. The advantage of the proposed scheme becomes more pronounced in user-overloaded regimes, indicating stronger robustness in challenging scenarios where other learning-based approaches often exhibit noticeable degradation. It also surpasses the WMMSE benchmark in underloaded systems and even in overloaded systems when the overloading factor is below certain threshold. These gains are achieved with fast inference and a model that is substantially lighter than the wireless foundation models.
\end{itemize}

\section{System Description and Problem Statement} \label{sec_sys_formu}

In this section, we introduce the downlink beamforming systems under both the Gaussian channel and the sparse channel assumptions, and formulate the MISO downlink beamforming problem within the \ac{L2O} framework.

\subsection{System Description} \label{sys_model}

We consider a single-cell downlink \ac{MISO} beamforming system, where a \ac{BS} is equipped with an antenna array of size $\bar{N}$. The maximum number of single-antenna users supported within the cell is denoted by $\bar{K}$. We assume that CSI is perfectly known at the \ac{BS}. In practice, users may dynamically enter and leave the serving cell, such that the instantaneous number of active users satisfies $K \leq \bar{K}$. Moreover, energy-efficient antenna selection strategies can be employed at the \ac{BS} to reduce power consumption~\cite{rajapaksha2023minimizing}, resulting in an instantaneous number of activated antennas $N \leq \bar{N}$. Consequently, the channel matrix between the activated antennas and the served users is denoted by $\tilde{\bm{H}} \triangleq [\tilde{\bm{h}}_1,\cdots,\tilde{\bm{h}}_K] \in \mathbb{C}^{N \times K}$, where $\tilde{\bm{h}}_k \in \mathbb{C}^{N}$ represents the channel between the \ac{BS} and user $k$. In this work, we consider two channel models. One is the Gaussian channel $\tilde{\bm{H}}$, whose entries are independently drawn as 
\begin{align} \label{gauss_chan_model}
[\tilde{\bm{H}}]_{ij} \overset{\text{i.i.d.}}{\sim} \mathcal{CN}\left(0, 1\right),
\end{align}
and hence $\mathbb{E}[\|\tilde{\bm{H}}\|_F^2]=KN$. Another case considers a sparse channel $\tilde{\bm{H}}$, where a BS equipped with $N$ activated antennas employs a \ac{ULA} with half-wavelength spacing to serve $K$ users, which is modeled as
\begin{align}  \label{sparse_chan_model}
\tilde{\bm{h}}_k = \sqrt{\frac{N}{L_p}} \sum_{\ell=1}^{L_p} \alpha_{k,\ell}  \left[ 1,e^{-j \pi \sin{\gamma_{k,\ell}}}, \dots, e^{-j \pi (N-1) \sin{\gamma_{k,\ell}}} \right]^T,
\end{align}
where $L_P$, $\alpha_{k,\ell} \sim \mathcal{CN}(0,1)$, and $\gamma_{k,\ell} \in [-\frac{\pi}{2},\frac{\pi}{2}]$ denote the number of scattering paths, the complex gain and the \ac{AOD} of the $\ell^{\text{th}}$ scattering path between the \ac{BS} and user $k$, respectively. The constant $\sqrt{\frac{N}{L_p}}$ in \eqref{sparse_chan_model} leads to $\mathbb{E}[\|\tilde{\bm{H}}\|_F^2]=KN$, which is consistent with the Gaussian channel case.


Suppose that the \ac{BS} applies a beamformer vector $\bm{w}_k \in \mathbb{C}^{N}$ to transmit the data symbol $x_k \in \mathbb{C}$ intended for user $k$. The received signal at user $k$ can then be expressed as
\begin{align} \label{receive_miso_full_digit}
y_k = \tilde{\bm{h}}_k^H \sum_{i=1}^{K} \bm{w}_i x_i + n_k, 
\end{align}
where $n_k \sim \mathcal{CN}(0,\sigma_k^2)$ denotes the additive noise at user $k$. Define the normalized channel matrix as $\bm{H}=[\bm{h}_1,\dots,\bm{h}_K]$, where $\bm{h}_k \triangleq \frac{\tilde{\bm{h}}_k}{\sigma_k}$, for $k \in [K]$. Denote the beamforming matrix as $\bm{W}=[\bm{w}_1,\dots,\bm{w}_K] \in \mathbb{C}^{N \times K}$. The achievable sum rate in this \ac{MISO} setting is given by
\begin{align} \label{sinr_rate_miso_full}
R_\text{sum}(\bm{H},\bm{W}) = \sum_{k=1}^K \log_2\left(1+\frac{\left|\bm{h}_k^H \bm{w}_k\right|^2}{1+\sum_{i \neq k} \left|\bm{h}_k^H \bm{w}_i\right|^2}\right).
\end{align}
The system is underloaded if $K < N$, and it is overloaded if $K \geq N$. In general, achieving high-rate beamformer solutions in the overloaded systems is more challenging than in the underloaded systems~\cite{joudeh2017rate}, due to a more delicate structure of \ac{IUI} in this regime.

\subsection{Problem Statement} \label{prob_formu}


Assume that the normalized channel follows a given distribution, i.e., $\bm{H} \sim p_H$. Our goal is to learn the optimal beamforming mapping:
\begin{align} \label{gauss_chan_prob_family}
\bm{W}^*(\bm{H}) = \arg\max_{\bm{W} \in \mathbb{C}^{N \times K}, \|\bm{W}\|_F^2 \leq P}  R_\text{sum}(\bm{H},\bm{W}),
\end{align}
for all dimensions $K \in [\bar{K}]$, $N \in [\bar{N}]$, and all $N \times K$ normalized channel realizations $\bm{H} \sim p_H$. In theory, such a mapping can be represented by a neural network $g_{\bm{\theta}}$, parameterized by $\bm{\theta}$, i.e., $\bm{W}=g_{\bm{\theta}}(\bm{H})$. However, conventional architectures of $g_{\bm{\theta}}$ fail to tackle this challenging learning problem. In particular, fully-connected networks, CNNs and RNNs exhibit poor convergence performance and need to be retrained when the system configuration changes, while GNNs also incur performance degradations in large-scale systems due to limited expressive power.  


Considering the difficulty of learning the direct mapping, we take the \ac{L2O} approach in \cite{johnston2023curriculum} to produce the beamformer solutions, which seeks to approximate the solution mapping by applying a $T$-layer neural network.
The parameters of the optimizer networks are allowed to vary across layers, which enhances the model expressive power. In particular, let $\mathcal{F}_{\bm{\theta}_t}$ denote the optimizer network at the $t^{\text{th}}$ layer, parameterized by $\bm{\theta}_t$, and let $\bm{W}^{(t)}$ denote the corresponding beamformer solution at the $t^{\text{th}}$ layer, for $t \in [T]$. To encourage performance improvement at every optimization step rather than focusing solely on the final objective value, we maximize the cumulative sum-rate along the optimization trajectory. To this end, the optimization problem family in \eqref{gauss_chan_prob_family} is reformulated as the following amortized learning problem:
\begin{align} \label{general_formu}
&\max_{\{\bm{\theta}_t\}_{t=1}^T} \ \mathbb{E}_{\bm{H} \sim p_H} \left[ \sum_{t=1}^T R_{\text{sum}}\left(\bm{H},\bm{W}^{(t)} \right) \right], \notag \\
&\text{s.t.} \ \ \|\bm{W}^{(t)}\|_F^2 \leq P, \ t \in [T], \ \forall (K,N) \in [\bar{K}] \times [\bar{N}],
\end{align}
where $\bm{H},\bm{W}^{(t)} \in \mathbb{C}^{N \times K}$. By learning the optimizer network parameters $\{\bm{\theta}_t\}_{t=1}^T$, the cost of the optimization is amortized over the training distribution, shifting the computational burden from online optimization in \eqref{gauss_chan_prob_family} to offline learning in \eqref{general_formu}. Nevertheless, solving \eqref{general_formu} remains challenging for the following two reasons: (a) non-convex sum-rate maximization in large-scale systems is inherently difficult and lacks a general-purpose solver, and (b) the optimizer network is required to generalize across optimizees with different system configurations without retraining, including the overloaded cases which are intrinsically hard to optimize. In the next section, we propose several novel architectural designs and training strategies to address these challenges.

\section{SALLO-M Transformer Model for Beamforming Optimization} \label{sec_bf_opt}

\subsection{SALLO Framework} \label{sa_l2o_frame}

\begin{figure*} [htbp]
    \centering
    \includegraphics[width=0.9\linewidth]{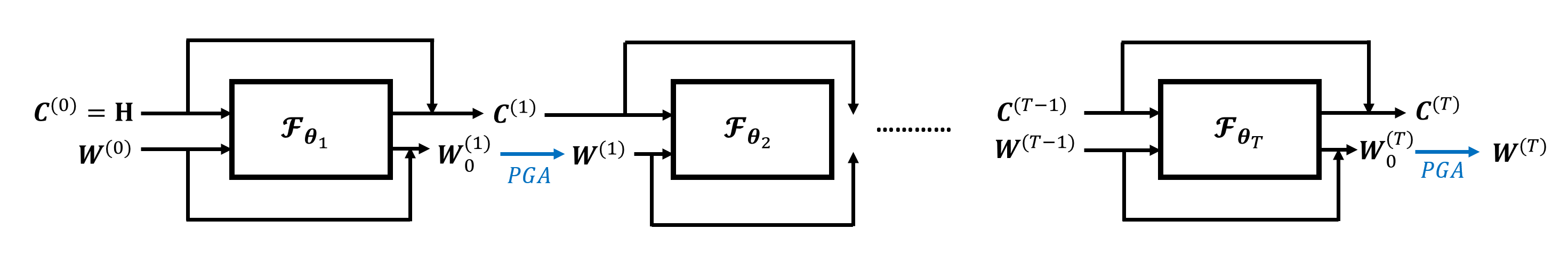}
    \caption{SALLO framework for beamformer optimization.}
    \label{L2O_framework}
\end{figure*}

In this section, we propose a \ac{SALLO} framework tailored to the formulation in \eqref{general_formu}, as illustrated in Fig.~\ref{L2O_framework}. For clarity, $\{\bm{C}^{(t)}\}_{t=0}^T$ and $\{\bm{W}^{(t)}\}_{t=0}^T$ in Fig.~\ref{L2O_framework} are referred to as the \emph{auxiliary variables} and \emph{beamformer solutions} across layers, respectively. It is observed that the auxiliary variable and the beamformer solution are jointly fed into the optimizer network and updated at each layer, which is different from the prior \ac{L2O} design that only updates the beamformer solution across layers \cite{johnston2023curriculum}. As a result, our design enlarges the effective feature space for optimization by updating the auxiliary variables at all layers.
Such a representation lifting enhances the model expressivity and facilitates the discovery of higher-quality solutions \cite{carreira2014distributed}. In addition, following \cite{he2016deep}, optimizing residual mappings rather than the original unreferenced mappings facilitates stable training of deep networks. Therefore, at the $t^{\text{th}}$ layer in Fig.~\ref{L2O_framework}, the optimizer network $\mathcal{F}_{\bm{\theta}_t}(\bm{C}^{(t-1)}, \bm{W}^{(t-1)})$ is trained to approximate the residuals $\Delta \bm{W}_0^{(t)} \triangleq \bm{W}_0^{(t)} - \bm{W}^{(t-1)}$ and $\Delta \bm{C}^{(t)} \triangleq \bm{C}^{(t)} - \bm{C}^{(t-1)}$, i.e.,
\begin{align} \label{residual_tf_net_map}
\Delta \bm{C}^{(t)}, \Delta \bm{W}_0^{(t)} = \mathcal{F}_{\bm{\theta}_t}(\bm{C}^{(t-1)}, \bm{W}^{(t-1)}).
\end{align}
In particular, the initial auxiliary variable is set as $\bm{C}^{(0)} = \bm{H}$, while the initial beamformer solution is set as the corresponding LMMSE beamformer, i.e.,
\begin{align} \label{mmse_bf}
&\bm{W}^{\text{m}}(\bm{H}) = [\bm{w}^{\text{m}}_1,\dots,\bm{w}^{\text{m}}_K], \notag \\
&\bm{w}^{\text{m}}_k = \sqrt{\frac{P}{K}} \cdot \frac{ (\bm{I}_N + \sum_{i=1}^{K} \frac{P}{K} \bm{h}_i \bm{h}_i^H)^{-1} \bm{h}_k }{ \Vert (\bm{I}_N + \sum_{i=1}^{K} \frac{P}{K} \bm{h}_i \bm{h}_i^H)^{-1} \bm{h}_k \Vert_2}, k \in [K].
\end{align} 

We further adopt the semi-amortized learning paradigm \cite{kim2018semi}, whereby each optimizer network is followed by a small number of gradient-ascent refinement steps to further improve the beamformer solution. Specifically, letting $\bm{W}_0^{(t)} \triangleq \Delta \bm{W}_0^{(t)} + \bm{W}^{(t-1)}$ denote the beamformer directly output by the optimizer network at the $t^{\text{th}}$ layer, the refined beamformer is then given by $\bm{W}^{(t)} \triangleq \bm{W}_Q^{(t)} =\mathcal{G}_Q(\bm{W}_0^{(t)})$, where $\mathcal{G}_Q$ denotes $Q$ steps of gradient ascent operations, given by
\begin{align} \label{gd_part_train}
\bm{W}_q^{(t)} &= \bm{W}_{q-1}^{(t)} + \eta_{w} \cdot \nabla_{\bm{W}} R_{\text{sum}}(\bm{H},\bm{W}_{q-1}^{(t)}), \notag \\
\bm{W}_q^{(t)} &= \sqrt{P} \cdot \frac{\bm{W}_q^{(t)}}{\|\bm{W}_q^{(t)}\|_F}, \ q \in [Q].
\end{align}

\begin{remark} \label{remark_1}
A special case of the above \ac{SALLO} framework is that the auxiliary variable $\bm{C}^{(t)}$ remains fixed, i.e., $\bm{C}^{(t)}=\bm{H}$, for $t=0,\dots,T$. Then only the beamformer solution is updated at the $t^{\text{th}}$ layer, i.e.,
\begin{align} \label{update_cons_H}
\Delta \bm{W}_0^{(t)} = \mathcal{F}_{\bm{\theta}_t}(\bm{H}, \bm{W}^{(t-1)}), \ 
\bm{W}^{(t)} = \mathcal{G}_Q(\bm{W}^{(t-1)}+\Delta \bm{W}_0^{(t)}).
\end{align}
This setting reduces the network size and the computational burden, at the expense of a reduced model expressivity, since the inputs to all layers are confined to a lower-dimensional feature space due to the fixed auxiliary variable. As a result, we refer to the simplified model in \eqref{update_cons_H} as the \ac{SALO} model, where the representation lifting is not applied.
\end{remark}


\subsection{Masked Transformer Architecture} \label{net_arch}

To achieve near-optimal beamformer solutions, we adopt Transformers to parameterize the optimizer networks $\{\mathcal{F}_{\bm{\theta}_t}\}_{t=1}^T$ in Fig.~\ref{L2O_framework}, motivated by their strong scalability to large-scale systems and state-of-the-art expressive power. In addition, the residual connections inherent in Transformer architectures facilitate stable training of deep networks, which aligns well with the proposed framework in Fig.~\ref{L2O_framework}. 

In this work, we further aim to learn a single Transformer model with \textbf{fixed} input and output dimensions that can produce high-quality beamformer solutions across varying user-antenna configurations. This objective poses a significant challenge for vanilla Transformers, which do not inherently exhibit a two-dimensional generalizability \cite{zhou2023algorithms,duan2025learning}. To address this problem, inspired by prior successful applications of masking operations \cite{cheng2022masked, fan2021mask}, we propose a masked Transformer architecture tailored to dynamic downlink systems. The key idea is to apply a structured scheduling of antenna-user masking during training, enabling a single model to be trained on channel samples with heterogeneous system configurations. This strategy allows the model to capture a more complex underlying data distribution and substantially improves its generalizability. Henceforth we will call our SALLO masked Transformer model the \textbf{SALLO-M Transformer model}.

\begin{figure*} [htbp]
    \centering
    \includegraphics[width=0.97\linewidth]{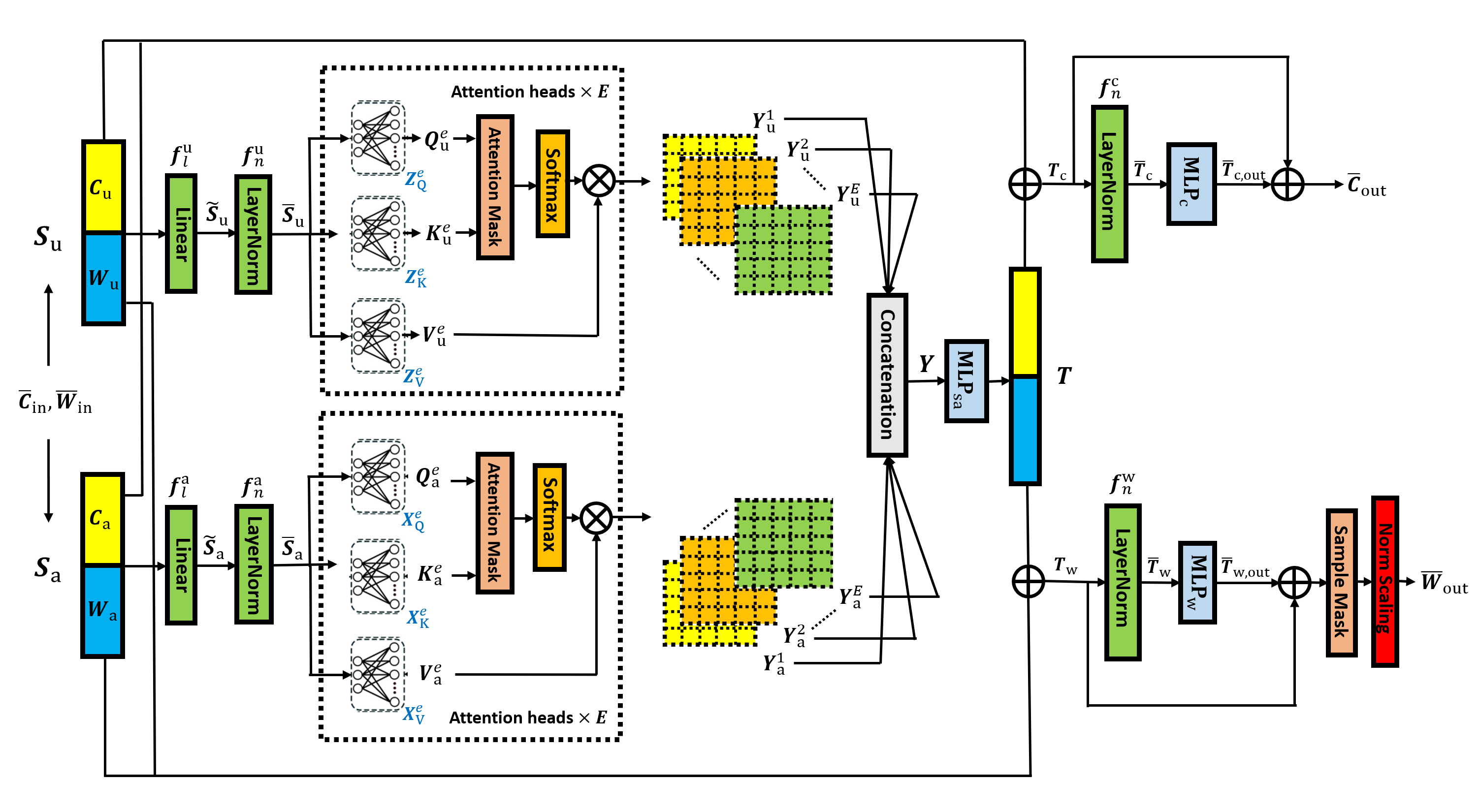}
    \caption{The architecture of each SALLO-M Transformer layer.}
    \label{arch_transformer}
\end{figure*}

The architecture of each layer of the SALLO-M Transformer model is illustrated in Fig.~\ref{arch_transformer}.\footnote{Note that in Fig.~\ref{arch_transformer}, a ``LayerNorm'' module has learnable shift and scale parameters for every feature dimension. In contrast, a ``Norm Scaling'' module simply scales the output beamformer to satisfy a fixed power constraint, hence has no learnable parameters.} The key feature is that the proposed model can naturally accommodate varying activated antenna-user configurations through the joint use of adaptive sample masking and attention masking strategies. The tunable parameters in each layer (i.e., $\bm{\theta}_t$) include the parameters of the two linear networks, the four Layer-Norm networks, the $6E$ attention weight matrices, and the three MLP networks shown in Fig.~\ref{arch_transformer}. 

\subsubsection{Sample Masking} \label{sample_mask_acquire}

Recall that $\bar{K}$ and $\bar{N}$ represent the maximum numbers of user and antenna, respectively. For a given user-antenna configuration with $K$ users and $N$ antennas, denote $\mathcal{R}_\text{u} \subseteq [\bar{K}]$ and $\mathcal{R}_\text{a} \subseteq [\bar{N}]$ as the index sets of the masked (out-of-cell) users and the masked (deactivated) antennas, respectively. Note that $|\mathcal{R}_\text{u}|=\bar{K}-K$ and $|\mathcal{R}_\text{a}|=\bar{N}-N$. For Gaussian channels, $\mathcal{R}_\text{u}$ and $\mathcal{R}_\text{a}$ can be randomly selected. On the other hand, for sparse channels, $\mathcal{R}_\text{u}$ is still randomly selected, but the activated antennas, i.e., the complement of $\mathcal{R}_\text{a}$, should be chosen as a contiguous subset of $N$ antennas, due to the \ac{ULA} structural constraint at the BS.

To produce a learning model that operates under various SNR values, we create a set $\mathcal{V}$ of noise variances corresponding to various SNRs. Then to obtain a normalized channel sample $\bm{H}^{(j)} \triangleq [\bm{h}^{(j)}_1,\dots,\bm{h}^{(j)}_K] \in \mathbb{C}^{N \times K}$, for a given configuration $(K,N)$, we generate a channel vector sample $\tilde{\bm{h}}_k^{(j)} \in \mathbb{C}^{N}$ according to \eqref{gauss_chan_model} for the Gaussian channel model or according to \eqref{sparse_chan_model} for the sparse channel model, and randomly pick a noise variance $(\sigma_k^2)^{(j)}$ from $\mathcal{V}$, and form $\bm{h}_k^{(j)} = \frac{\tilde{\bm{h}}_k^{(j)}}{(\sigma_k)^{(j)}} \in \mathbb{C}^{N}$, $k \in [K]$. As a result, a normalized channel sample $\bm{H}^{(j)}$ for a certain SNR is obtained. Then we compute the corresponding LMMSE beamformer $\bm{W}^{\text{m}}(\bm{H}^{(j)}) \in \mathbb{C}^{N \times K}$ in \eqref{mmse_bf}. Finally, based on $\mathcal{R}_\text{u}$ and $\mathcal{R}_\text{a}$, we perform zero-padding on $\bm{H}^{(j)}$ and $\bm{W}^{\text{m}}(\bm{H}^{(j)})$ to obtain the full-size matrices $\bar{\bm{C}}^{(0)} \in \mathbb{C}^{\bar{N} \times \bar{K}}$ and $\bar{\bm{W}}^{(0)} \in \mathbb{C}^{\bar{N} \times \bar{K}}$, which constitute the inputs to the first layer.   


\subsubsection{Token Sequence Construction} \label{input_token}
Inspired by the dual-path strategy in \cite{geng2021deep}, to exploit the two-dimensional structure of the channel matrix, we adopt a \textbf{dual tokenization} strategy and construct two complementary token sequences for each layer input: a user-level sequence and an antenna-level sequence. The former emphasizes inter-user coupling that is closely related to the interference structure, while the latter captures spatial correlations across transmit antennas. In addition, this construction treats variations in the numbers of users and antennas in a symmetric manner, since both can be handled through sequence-level masking.

We then construct these two token sequences based on the inputs to each layer. Let the full-size matrices $(\bar{\bm{C}}_{\text{in}},\bar{\bm{W}}_{\text{in}})$ denote the input auxiliary variable and the input beamformer solution to a certain layer, respectively. Each of $\bar{\bm{C}}_{\text{in}} \in \mathbb{C}^{\bar{N} \times \bar{K}}$ and $\bar{\bm{W}}_{\text{in}} \in \mathbb{C}^{\bar{N} \times \bar{K}}$ can be interpreted as a sequence of user-level vectors, i.e., $\bar{\bm{C}}_{\text{in}} =[\bm{c}_1,\dots,\bm{c}_{\bar{K}}]$ and $\bar{\bm{W}}_{\text{in}}=[\bm{w}_1,\dots,\bm{w}_{\bar{K}}]$. For each user-level vector $\bm{c}_k \in \mathbb{C}^{\bar{N}}$, the real and imaginary components are separated and concatenated as $\bm{c}^{(\text{u})}_k = [\Re (\bm{c}_k);\Im (\bm{c}_k)] \in \mathbb{R}^{2\bar{N}}$, yielding the real-valued user-level matrix $\bm{C}_\text{u} = [\bm{c}^{(\text{u})}_1,\dots,\bm{c}^{(\text{u})}_{\bar{K}}] \in \mathbb{R}^{2\bar{N} \times \bar{K}}$. Similarly, for each user-level beamformer vector $\bm{w}_k \in \mathbb{C}^{\bar{N}}$, we obtain $\bm{w}^{(\text{u})}_k = [\Re(\bm{w}_k);\Im(\bm{w}_k)] \in \mathbb{R}^{2\bar{N}}$, and form the real-valued user-level beamformer matrix $\bm{W}_\text{u} = [\bm{w}^{(\text{u})}_1,\dots,\bm{w}^{(\text{u})}_{\bar{K}}] \in \mathbb{R}^{2\bar{N} \times \bar{K}}$. Finally, the user-level token sequence can be formed as $\bm{S}_\text{u} \triangleq [\bm{C}_\text{u},\bm{W}_\text{u}] \in \mathbb{R}^{2\bar{N} \times 2\bar{K}}$.

Likewise, the transposed inputs $\bar{\bm{C}}_{\text{in}}^T \in \mathbb{C}^{\bar{K} \times \bar{N}}$ and $\bar{\bm{W}}_{\text{in}}^T \in \mathbb{C}^{\bar{K} \times \bar{N}}$ can be viewed as sequences of antenna-level vectors. Similar to the user-level token constructions, we obtain the real-valued matrices $\bm{C}_\text{a} \in \mathbb{R}^{2\bar{K} \times \bar{N}}$ and $\bm{W}_\text{a} \in \mathbb{R}^{2\bar{K} \times \bar{N}}$, hence the antenna-level token sequence can be formed as $\bm{S}_\text{a} \triangleq [\bm{C}_\text{a},\bm{W}_\text{a}] \in \mathbb{R}^{2\bar{K} \times 2\bar{N}}$. For notational convenience, we set $\bar{K} = \bar{N} = L$ throughout this paper. Consequently, each token has a feature dimension $2L$, and the corresponding sequence length is also $2L$.


Next, the sequences $\bm{S}_\text{u}$ and $\bm{S}_\text{a}$ are processed independently by two embedding layers. Each embedding layer consists of a \ac{FC} layer followed by a \ac{TN} layer. Given input tokens of feature dimension $2L$, the \ac{FC} layer projects them into an $M$-dimensional space, yielding $\tilde{\bm{S}}_\text{u} \in \mathbb{R}^{M \times 2L}$ and $\tilde{\bm{S}}_\text{a} \in \mathbb{R}^{M \times 2L}$. Typically, the original tokens are projected into a larger feature space to enhance the representation capability, i.e., $M > 2L$. The subsequent \ac{TN} layers normalize each token independently, producing the normalized embeddings $\bar{\bm{S}}_\text{u} \in \mathbb{R}^{M \times 2L}$ and $\bar{\bm{S}}_\text{a} \in \mathbb{R}^{M \times 2L}$. Importantly, no positional encoding is employed, as it would break the inherent permutation equivariance of the Transformer \cite{li2024hpe,duan2025learning}, which is essential for modeling the unordered user-antenna structure and ensuring generalizability across varying system configurations.



\subsubsection{Multi-head Self-attention with Attention Masking} \label{MHSA_structure}

The \ac{MHSA} scheme employs multiple attention heads to project the embedded tokens into distinct subspaces, thereby capturing diverse interaction patterns while improving training stability. Denote the head dimension and the number of attention heads as $D_e$ and $E$, respectively.\footnote{The head dimension $D_e$ is typically fixed at 64 or 128.} Specifically, for the $e^{\text{th}}$ attention head, the query, key, and value matrices of dimension $2L \times D_e$ corresponding to the token sequence $\bar{\bm{S}}_\text{u}$ are obtained as
\begin{align} \label{qkv_s_seq}
\bm{Q}^e_\text{u} = \bar{\bm{S}}_\text{u}^T \bm{Z}_Q^e, \ \bm{K}^e_\text{u} = \bar{\bm{S}}_\text{u}^T \bm{Z}_K^e, \ \bm{V}^e_\text{u} = \bar{\bm{S}}_\text{u}^T \bm{Z}_V^e, \ e \in [E],
\end{align}
where $\bm{Z}_Q^e, \bm{Z}_K^e, \bm{Z}_V^e \in \mathbb{R}^{M \times D_e}$ are the learnable query, key, and value weight matrices, respectively. Similarly, for token sequence $\bar{\bm{S}}_\text{a}$, the query, key and value matrices of dimension $2L \times D_e$ are given by
\begin{align} \label{qkv_t_seq}
\bm{Q}^e_\text{a} = \bar{\bm{S}}_\text{a}^T \bm{X}_Q^e, \ \bm{K}^e_\text{a} = \bar{\bm{S}}_\text{a}^T \bm{X}_K^e, \ \bm{V}^e_\text{a} = \bar{\bm{S}}_\text{a}^T \bm{X}_V^e, \ e \in [E],
\end{align}
with $\bm{X}_Q^e, \bm{X}_K^e, \bm{X}_V^e \in \mathbb{R}^{M \times D_e}$ being the corresponding learnable weight matrices. 


To prevent attention from being diffused over masked users or antennas, we construct the attention masking matrices $\bm{M}_\text{U} \in \mathbb{R}^{2L \times 2L}$ and $\bm{M}_\text{A} \in \mathbb{R}^{2L \times 2L}$ based on the masking index sets $\mathcal{R}_\text{u}$ and $\mathcal{R}_\text{a}$, given as follows, 
\begin{align} \label{attn_mask_mat_def}
\bm{M}_\text{U} = [\bm{M}_\text{u},\bm{M}_\text{u};\bm{M}_\text{u},\bm{M}_\text{u}], \ \bm{M}_\text{A} = [\bm{M}_\text{a},\bm{M}_\text{a};\bm{M}_\text{a},\bm{M}_\text{a}],
\end{align}
where 
\begin{align} \label{attn_mask_user}
\bm{M}_\text{u}(i,j) =  
\begin{cases} 
-\infty, & \ i \in \mathcal{R}_\text{u} \ \text{or} \ j \in \mathcal{R}_\text{u}, \\
0, & \ \text{otherwise},
\end{cases} 
\end{align}
and
\begin{align} \label{attn_mask_ant}
\bm{M}_\text{a}(i,j) =  
\begin{cases} 
-\infty, & \ i \in \mathcal{R}_\text{a} \ \text{or} \ j \in \mathcal{R}_\text{a}, \\
0, & \ \text{otherwise}.
\end{cases} 
\end{align}
Note that the block-wise partitioning in \eqref{attn_mask_mat_def} arises from the joint inclusion of channel vectors and beamformer vectors in the token construction. These masks are directly added to the user-level and antenna-level attention score matrices of dimension $2L \times 2L$ (i.e., the scaled product of query and key matrices), given as follows,
\begin{align} \label{mask_attn_output}
\bm{Y}_\text{u}^e = \text{softmax} \left(\frac{\bm{Q}_\text{u}^e (\bm{K}_\text{u}^e)^T}{\sqrt{D_e}} + \bm{M}_\text{U} \right) \bm{V}_\text{u}^e, \ e \in [E], \notag \\
\bm{Y}_\text{a}^e = \text{softmax} \left(\frac{\bm{Q}_\text{a}^e (\bm{K}_\text{a}^e)^T}{\sqrt{D_e}} + \bm{M}_\text{A} \right) \bm{V}_\text{a}^e, \ e \in [E],
\end{align}
such that all attention scores associated with masked users or masked antennas are assigned $-\infty$ and are therefore nullified after the Softmax operation. 

The attention outputs in \eqref{mask_attn_output} are then concatenated along the feature dimension to form
\begin{align} \label{combine_y}
\bm{Y}=[\bm{Y}_\text{u}^1,\dots,\bm{Y}_\text{u}^E,\bm{Y}_\text{a}^1,\dots,\bm{Y}_\text{a}^E]^T \in \mathbb{R}^{2D \times 2L},
\end{align}
where $D \triangleq D_e E$ is known as the model dimension of a Transformer network.\footnote{Typically, the model dimension $D$ is around 500 to 1000, much larger than the original token dimension $2L$.} To effectively aggregate the attention outputs from all attention heads, we apply a \ac{MLP} after the concatenation in \eqref{combine_y}, denoted by $\text{MLP}_\text{sa}(\cdot): \mathbb{R}^{2D} \rightarrow \mathbb{R}^{2L}$, which produces $\bm{T} = \text{MLP}_\text{sa}(\bm{Y}) \in \mathbb{R}^{2L \times 2L}$. The \ac{MLP} typically consists of \ac{FC} layers and GELU activation layers, thereby enhancing the expressivity by introducing non-linearity into multi-head aggregation process. To enable the residual mapping learning in \eqref{residual_tf_net_map}, the input real-valued auxiliary variables and beamformer solutions are directly added to the outputs of the \ac{MHSA} component, to obtain
\begin{align} \label{mhsa_resi_output}
\bm{T}_\text{c} &= \bm{T}[:,1:L] + \bm{S}_\text{u}[:,1:L] + \bm{S}_\text{a}[:,1:L], \notag \\
\bm{T}_\text{w} &= \bm{T}[:,L+1:2L] + \bm{S}_\text{u}[:,L+1:2L] + \bm{S}_\text{a}[:,L+1:2L].
\end{align}


\subsubsection{Output MLP} \label{out_mlp}

This component outputs the updated auxiliary variable and beamformer solution via two dedicated \ac{MLP} networks, denoted by $\text{MLP}_\text{c}(\cdot): \mathbb{R}^{2L} \rightarrow \mathbb{R}^{2L}$ and $\text{MLP}_\text{w}(\cdot): \mathbb{R}^{2L} \rightarrow \mathbb{R}^{2L}$, respectively. Specifically, the intermediate results in \eqref{mhsa_resi_output} are first normalized by \ac{TN} layers, yielding $\bar{\bm{T}}_\text{c} \in \mathbb{R}^{2L \times L}$ and $\bar{\bm{T}}_\text{w} \in \mathbb{R}^{2L \times L}$. The normalized results are then fed into the corresponding \ac{MLP} networks:
\begin{align} \label{mlp_h_w_out}
\bar{\bm{T}}_{\text{c,out}} = \text{MLP}_\text{c}(\bar{\bm{T}}_{\text{c}}), \ \bar{\bm{T}}_{\text{w,out}} = \text{MLP}_\text{w}(\bar{\bm{T}}_{\text{w}}).
\end{align}
Similar to the residual connections in \eqref{mhsa_resi_output}, the outputs of the MLP components are given by
\begin{align} \label{output_mlp_resi}
\bm{T}_{\text{c,out}} = \bar{\bm{T}}_{\text{c,out}} + \bm{T}_{\text{c}}, \ \bm{T}_{\text{w,out}} = \bar{\bm{T}}_{\text{w,out}} + \bm{T}_{\text{w}}.
\end{align}
We then combine the real and imaginary parts to recover the complex-valued matrices:
\begin{align} \label{construct_hw_out}
\bar{\bm{C}}_{\text{out}} &= \bm{T}_{\text{c,out}}[1:L]+j \cdot \bm{T}_{\text{c,out}}[L+1:2L], \notag \\
\bar{\bm{W}}_{\text{out}} &= \bm{T}_{\text{w,out}}[1:L]+j \cdot \bm{T}_{\text{w,out}}[L+1:2L].
\end{align}
To enforce the inactivity of masked users and antennas in the updated beamformer solution, based on $\mathcal{R}_\text{u}$ and $\mathcal{R}_\text{a}$, we set the corresponding entries in $\bar{\bm{W}}_{\text{out}}$ to zero and then normalize $\bar{\bm{W}}_{\text{out}}$ to satisfy the transmit power constraint $\|\bar{\bm{W}}_{\text{out}}\|^2 = P$. As a result, the masked Transformer at the $t^{\text{th}}$ layer maps the input pair $(\bar{\bm{C}}_{\text{in}},\bar{\bm{W}}_{\text{in}})=(\bar{\bm{C}}^{(t-1)},\bar{\bm{W}}^{(t-1)})$ to the output pair $(\bar{\bm{C}}_{\text{out}},\bar{\bm{W}}_{\text{out}})=(\bar{\bm{C}}^{(t)},\bar{\bm{W}}_0^{(t)})$.



Then we remove zero entries in $\bar{\bm{W}}_0^{(t)} \in \mathbb{C}^{\bar{N} \times \bar{K}}$ to obtain $\bm{W}^{(t)}_0 \in \mathbb{C}^{N \times K}$. Similar to \eqref{gd_part_train}, we further perform a $Q$-step gradient ascent refinement starting from $\bm{W}^{(t)}_0$ based on the activated true channel $\bm{H} \in \mathbb{C}^{N \times K}$, i.e.,
\begin{align} \label{gd_mask_arch}
\bm{W}_{q}^{(t)} &= \bm{W}_{q-1}^{(t)} + \eta_{w} \cdot \nabla_{\bm{W}} R_{\text{sum}}(\bm{H},\bm{W}_{q-1}^{(t)}), \notag \\
\bm{W}_{q}^{(t)} &= \sqrt{P} \cdot \frac{\bm{W}_{q}^{(t)}}{\|\bm{W}_{q}^{(t)}\|_F}, \ q \in [Q],
\end{align}
which outputs the refined solution $\bm{W}^{(t)} \triangleq \bm{W}_{Q}^{(t)}$. Finally, we perform zero-padding on $\bm{W}^{(t)} \in \mathbb{C}^{N \times K}$ to obtain the full-size beamformer $\bar{\bm{W}}^{(t)} \in \mathbb{C}^{\bar{N} \times \bar{K}}$, which constitutes the inputs to the next layer.


\begin{figure*}
    \centering
    \includegraphics[width=0.57\linewidth]{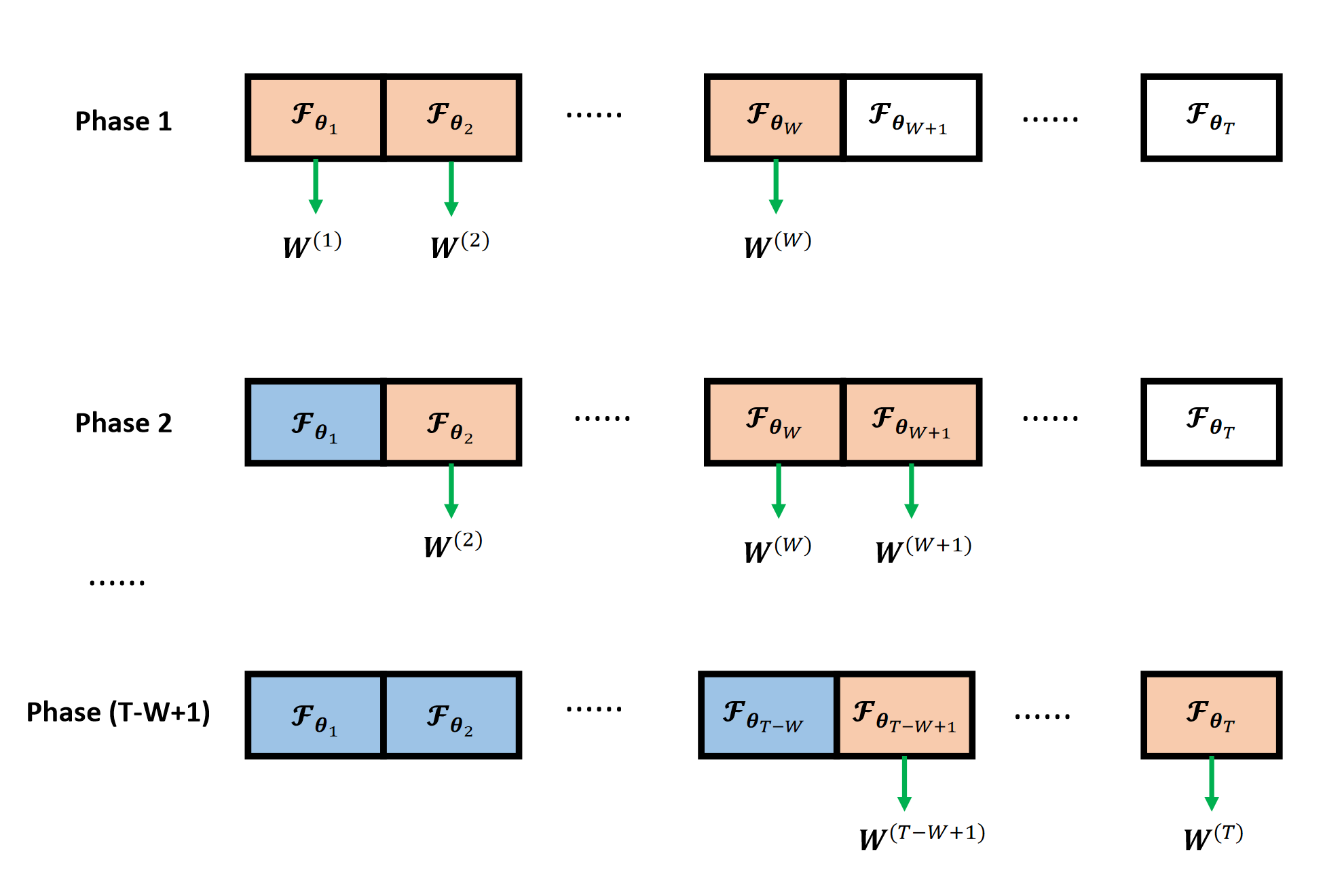}
    \caption{Sliding-window training of the proposed model. In each training epoch of a given phase, the forward pass covers both blue and orange blocks, while the back-propagation covers the orange blocks only.} 
    \label{slid_win_process}
\end{figure*}

\subsection{Enhanced Training Methods} \label{transformer_train_strategy}

In this section, we present several novel techniques to improve the training convergence of the proposed SALLO-M Transformer model.

\subsubsection{Sliding-window Training Method} \label{slide_window_method}

The end-to-end training of the model in Fig.~\ref{L2O_framework} becomes challenging when the total number of layers $T$ is large, as backpropagating through all layers leads to unstable gradients and high memory consumption. To address these issues and reduce the computational cost during training, we adopt a sliding-window strategy inspired by the layer-wise training method in \cite{bengio2006greedy}. Concretely, assume a window spanning layers $[t_s,t_e]$ ($W=t_e-t_s+1$ is the fixed window size) moves along the depth of the model, and the layers are partitioned into three disjoint regions accordingly: All networks preceding the $t_s^{\text{th}}$ layer are used in an inference-only manner, with parameters frozen; all networks inside layers $[t_s,t_e]$ are trainable and receive parameter updates by \ac{SGD}; and all layers after the $t_e^{\text{th}}$ layer are not yet included in the current network. The corresponding loss function is given by:
\begin{align} \label{sliding_loss}
L(\bm{\theta}_{t_s},\dots,\bm{\theta}_{t_e}) =  \mathbb{E}_{\bm{H} \sim p_H} \left[ \sum_{t=t_s}^{t_e} R_{\text{sum}}\left(\bm{H},\bm{W}^{(t)} \right) \right].
\end{align}
This design focuses optimization on a small subset of layers at each step, improving computational efficiency and stabilizing the optimization dynamics. The training process is illustrated in Fig.~\ref{slid_win_process}. As the window moves along, every Transformer layer is progressively updated, ensuring that the full network is eventually trained without suffering from the instability caused by excessively long training horizons. 



\subsubsection{Curriculum Learning with Random Masking} \label{cl_random_mask}




We employ a random masking strategy to generate a diverse training dataset, improving the model’s robustness to changes in both the numbers and identities of active users and antennas. To support this, we vary the system parameters $(K,N)$ during training and refer to each specific pair $(K,N)$ as a \emph{system configuration}. We then design a scheduling strategy over $(K,N)$ that acts as a sample-based curriculum learning (CL) scheme to enhance the effectiveness of this sliding-window training process. Specifically, \ac{CL} is a training paradigm in which neural networks are exposed to tasks of increasing difficulty, progressing from easier to harder ones. Such a strategy has been shown to mitigate premature convergence --- an issue that is especially pronounced in large-scale system optimization with highly non-convex landscapes --- by encouraging broader exploration and acting as an implicit regularizer \cite{bengio2009curriculum,johnston2023curriculum}. 

Following the principle of \ac{CL}, the model is initially trained on channel samples for which high-quality beamforming solutions are relatively easy to obtain, and the training progressively shifts toward more challenging channel realizations. As discussed in Sec.~\ref{sys_model}, beamforming optimization generally becomes more difficult as the number of users increases, since the resulting \ac{IUI} exhibits increasingly intricate coupling structures, thereby complicating effective interference suppression. Accordingly, we define a sequence of user numbers $\mathcal{K} = [K_1, \dots, K_{n_\text{u}}]$ and a sequence of antenna numbers $\mathcal{N} = [N_1, \dots, N_{n_\text{a}}]$, such that $0 < K_1 < \cdots < K_{n_\text{u}} = \bar{K}$ and $0 < N_1 < \cdots < N_{n_\text{a}} = \bar{N}$. For each configuration $(K,N) \in \mathcal{K} \times \mathcal{N}$,
the attention masking matrices are generated based on the randomly-sampled index sets $\mathcal{R}_{\text{u}}$ and $\mathcal{R}_{\text{a}}$, and the corresponding channel matrix $\bm{H}$ is generated according to either the Gaussian channel model in \eqref{gauss_chan_model} or the sparse channel model in \eqref{sparse_chan_model}. Although the training procedure does not exhaustively traverse all possible configurations (since $\mathcal{K} \subset [\bar{K}]$ and $\mathcal{N} \subset [\bar{N}]$), the resulting model is expected to generalize effectively to all configurations $(K,N) \in [\bar{K}] \times [\bar{N}]$ during inference.

\subsubsection{Sample Replay} \label{sam_replay}

A potential issue of the CL scheme described in Sec.~\ref{cl_random_mask} is \emph{model forgetting}, whereby performance on earlier configurations may deteriorate as the model adapts to later configurations in the training curriculum~\cite{toneva2018empirical}. To mitigate this effect, we incorporate a lightweight sample replay strategy~\cite{rolnick2019experience}. Specifically, during training at each configuration $(K, N)$, every mini-batch is composed of two parts: (i) the majority of samples drawn from the current configuration, and (ii) a small fraction of replay samples randomly selected from previous configurations. This replay mechanism helps preserve knowledge acquired in earlier stages, thereby maintaining performance consistency across stages and improving the overall robustness of the curriculum learning process.

Finally, the overall training pipeline of the proposed model is summarized in Algorithm~\ref{alg_1}. During inference, for a given channel sample $\bm{H}$ and the corresponding masking index sets $\mathcal{R}_\text{u}$ and $\mathcal{R}_\text{a}$, we compute the LMMSE beamformer, and form the inputs $(\bar{\bm{C}}^{(0)},\bar{\bm{W}}^{(0)})$. Then we perform a $T$-layer forward pass to obtain the final beamformer solution $\bm{W}^{(T)}$.

\setlength{\algorithmicindent}{2em}

\begin{algorithm} 
\caption{SALLO-M Transformer training for dynamic multi-user downlink beamforming} \label{alg_1}
\begin{algorithmic} [1]
\STATE \textbf{Parameters}: Number of layers $T$, sliding-window length $W$, activated user schedule $\mathcal{K}=[K_1,\dots,K_{n_\text{u}}]$, activated antenna schedule $\mathcal{N}=[N_1,\dots,N_{n_\text{a}}]$, training batch size $B_\text{t}$, replay sample number per training batch $B_\text{r}$
\STATE Randomly initialize model parameters $\{\bm{\theta}_t\}_{t=1}^{T}$, initialize the sliding-window as $[t_s,t_e] \leftarrow [1,W]$, and initialize the configuration as $(K,N) \leftarrow (K_1,N_1)$
\STATE \textbf{for} each sliding-window position \textbf{do}
\STATE \quad \textbf{for} each configuration $(K,N)$ \textbf{do}
\STATE \quad \hspace{0.4em} - Obtain a training batch $\{\bar{\bm
H}_j\}_{j=1}^{B_\text{t}}$, with $B_\text{t}-B_\text{r}$ \\
\quad \hspace{1em} channel samples of configuration $(K,N)$, and $B_\text{r}$ \\
\quad \hspace{1em} replay samples 
\STATE \quad \quad \textbf{for} $j=1,\dots,B_\text{t}$ \textbf{do}
\STATE \quad \quad \hspace{0.4em} - Initialize inputs $(\bar{\bm{C}}^{(0)},\bar{\bm{W}}^{(0)})$ based on $\bar{\bm{H}}_j$
\STATE \quad \quad \quad \textbf{for} $t=1,\dots,t_e$ \textbf{do} \quad \textit{// forward pass}
\STATE \quad \quad \quad \hspace{0.4em} - Obtain the network outputs in \eqref{construct_hw_out}
\STATE \quad \quad \quad \hspace{0.4em} - Mask, normalize, and refine the output \\
\STATE \quad \quad \quad \hspace{1em} beamformer, yielding $\bm{W}^{(t)}_j$ by \eqref{gd_mask_arch}
\STATE \quad \quad \quad \hspace{0.4em} - Feed $(\bar{\bm{C}}_j^{(t)},\bar{\bm{W}}_j^{(t)})$ to the next layer
\STATE \quad \quad \quad \textbf{end for}
\STATE \quad \quad \hspace{0.4em} - Obtain $\{\bm{W}_j^{(t)}\}_{t=t_s}^{t_e}$, compute the loss in \eqref{sliding_loss}
\STATE \quad \quad \textbf{end for}
\STATE \quad \hspace{0.4em} - Update $\{\bm{\theta}_t\}_{t=t_s}^{t_e}$ by SGD
\STATE \quad \textbf{end for}
\STATE \textbf{end for}
\STATE \textbf{Output:} Trained model parameters $\{\bm{\theta}^{*}_t\}_{t=1}^{T}$
\end{algorithmic}
\end{algorithm}

\begin{remark} \label{remark_2}
We now provide the overall \textbf{inference complexity} of the proposed SALLO-M Transformer model. For each Transformer layer, the complexities of the embedding modules, the \ac{MHSA} module, the MLP modules, and the gradient ascent module are $\mathcal{O}(L^2 M)$, $\mathcal{O}(LMD)$, $\mathcal{O}(L^2 D)$, and $\mathcal{O}(Q_i L^3)$, respectively. Consequently, the $T$-layer SALLO-M Transformer model has a complexity of $\mathcal{O}(T(L^2 M+LMD+L^2 D+Q_i L^3))$.  
\end{remark}

\section{Numerical Results} \label{simu_result}

\subsection{Network Parameters} \label{basic_setup}

We consider a downlink multi-user \ac{MISO} system, where a \ac{BS} equipped with $\bar{N}=40$ transmit antennas serves up to $\bar{K}=40$ single-antenna users within the cell. The Gaussian channel samples and the sparse channel samples are generated according to \eqref{gauss_chan_model} and \eqref{sparse_chan_model}, respectively. All users are assumed to experience identical noise variance, i.e., $\sigma_k^2 = \sigma^2$ for $\forall k \in [K]$. The transmit power is normalized such that the beamformer satisfies $\|\bm{W}\|_F^2 = P = 1$, and hence the corresponding \ac{SNR} is given as $\text{SNR} \triangleq \frac{1}{\sigma^2}$. 

The parameters in Algorithm~\ref{alg_1} are set as follows. The activated user and antenna schedules are set to $\mathcal{K}=[8,16,24,32,40]$ and $\mathcal{N}=[8,16,24,32,40]$, respectively. Each configuration $(K,N)$ spans $12$ training epochs, hence the sliding window is moved forward by one step every $|\mathcal{K}| \cdot |\mathcal{N}| \cdot 12=300$ training epochs. The channel SNR is sampled from the SNR set $\mathcal{V}_{\text{train}}=[5,10,15,20] \ \text{dB}$. The training batch size is $B_\text{t}=128$, and the replay sample number per training batch is $B_\text{r}=32$. Furthermore, to comprehensively evaluate the model's generalizability, we construct a test dataset of size $B_\text{i}=6.4 \times 10^5$, which contains $20$ test samples for each configuration $(K,N) \in [\bar{K}] \times [\bar{N}]$ and each SNR value in $\mathcal{V}_\text{test}=[2.5,5,\dots,17.5,20] \ \text{dB}$. The learning rate is initialized as $\eta = 1 \times 10^{-4}$ and decayed to $3\times10^{-5}$ following a cosine decay schedule. The step size of gradient ascent in \eqref{gd_part_train} is set to $\eta_\text{w}=10^{-2}$. The layer number of the SALLO-M Transformer model is set to $T=10$. 


Moreover, Table~\ref{tab_tf_net} summarizes the number of parameters of all components in a single Transformer block shown in Fig.~\ref{arch_transformer}. Two fully-connected (FC) layers $f^{\text{u}}_\ell$ and $f^{\text{a}}_\ell$ map tokens from dimension $2L=80$ to $M=128$, resulting in $80 \times 128 \times 2 = 20480$ parameters. For the LayerNorm (LN) modules, $(f^{\text{u}}_n,f^{\text{a}}_n)$ and $(f^{\text{c}}_n,f^{\text{w}}_n)$ normalize tokens with dimensions $M=128$ and $2L=80$, respectively. Since each LN learns the shift and scaling parameters for every feature dimension, the four LN modules together introduce $(128+80) \times 2 \times 2 = 832$ parameters. Each attention weight matrix maps a token of length $M=128$ to head dimension $D_e=64$. For each token sequence $\bm{S}_\text{u}$ and $\bm{S}_\text{a}$, the number of attention heads is $E=12$, and $3E$ such matrices are used in the attention operations, yielding $128 \times 64 \times 36 = 294912$ parameters per sequence. 
Moreover, the MLP that aggregates the outputs of all attention heads, denoted as $\text{MLP}_\text{sa}$ in Fig.~\ref{arch_transformer}, contains three FC layers with sizes $768 \times 160$, $160 \times 80$, and $80 \times 40$, resulting in $138880$ parameters. The other two MLP networks, $\text{MLP}_\text{c}$ and $\text{MLP}_\text{w}$, introduce $2 \times 38400 = 76800$ parameters. Overall, the proposed $10$-layer SALLO-M Transformer model contains approximately $8.27 \times 10^6$ parameters, which is significantly smaller than most existing foundation models (e.g., in \cite{xu2025llm,li2025bert4beam}) whose parameter size is at least $10^8 \sim 10^9$, and thus is suitable for deployment on resource-constrained devices.

\begin{table}[t]
\centering
\caption{Parameter sizes of all components in one Transformer block. }
\label{tab_tf_net}
\begin{tabular}{c c c}
\hline
\textbf{Name} & \textbf{Notations} & \textbf{Size} \\
\hline
FC layer & $f_\ell^\text{u}, f_\ell^\text{a}$ & $20480$ \\
LN layer & $f_n^\text{u}, f_n^\text{a}, f_n^\text{c}, f_n^\text{w}$ & $832$ \\
Weight matrices & $\{\bm{Q}_\text{u}^e,\bm{K}_\text{u}^e,\bm{V}_\text{u}^e\}_{e=1}^E$ & $294912$ \\
Weight matrices & $\{\bm{Q}_\text{a}^e,\bm{K}_\text{a}^e,\bm{V}_\text{a}^e\}_{e=1}^E$ & $294912$ \\
Attention MLP & $\text{MLP}_{\text{sa}}$ & $138880$ \\
Output MLP & $\text{MLP}_\text{c}, \text{MLP}_\text{w}$ & $76800$ \\
\hline
\end{tabular}
\end{table}

\subsection{Ablation Study} \label{ablation_results}

This subsection presents ablation studies on the key components of the proposed SALLO-M Transformer, including several architectural designs --- semi-amortized learning, representation lifting, and user–antenna dual tokenization --- and several training strategies, namely the sliding-window method, sample replay, and curriculum learning with random masking. The following two non-learning methods are used as baselines: 
\begin{itemize}
    \item \textbf{WMMSE}: The classical iterative beamforming optimization algorithm \cite{shi2011iteratively}.
    \item \textbf{LMMSE}: the suboptimal LMMSE beamformer given by \eqref{mmse_bf}.
\end{itemize}

\begin{figure} [htbp]
    \centering
    \includegraphics[width=1.05\linewidth]{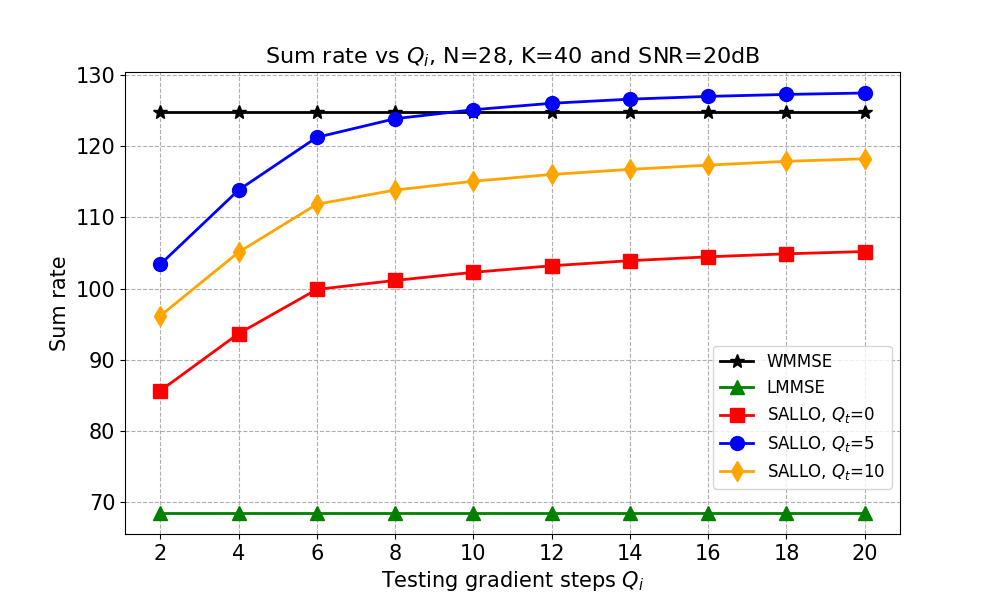}
    \caption{Performance under different numbers of gradient steps during training ($Q_t$) and testing ($Q_i$). }
    \label{perf_diff_gd_steps}
\end{figure}

We first evaluate the effectiveness of the proposed \textbf{semi-amortized learning} strategy by examining how the number of gradient ascent steps at each layer—denoted by $Q_t$ during training and $Q_i$ during inference—affects the final sum-rate performance. Specifically, three SALLO-M Transformer models are trained with $Q_t = 0, 5, 10$, respectively, and subsequently evaluated under Gaussian channels with $(K,N) = (40,28)$ and $\text{SNR}=20$ dB, while varying $Q_i$. Fig.~\ref{perf_diff_gd_steps} plots the achieved sum rate as a function of $Q_i$ for the three trained models. It can be observed that the model trained with $Q_t=5$ significantly outperforms the one trained with $Q_t=0$, demonstrating the effectiveness of the proposed semi-amortized design in producing high-quality beamformer solutions. Interestingly, the model with $Q_t=5$ also outperforms that with $Q_t=10$, indicating that an excessively large $Q_t$ leads to inferior performance. During inference, employing a larger number of gradient ascent steps, i.e., $Q_i > Q_t$, further improves the achievable sum rate. Based on the results in Fig.~\ref{perf_diff_gd_steps}, we set $Q_t=5$ and $Q_i=10$ for all subsequent simulations, which strikes a favorable balance between performance and computational complexity.

\begin{figure} [htbp]
    \centering
    \includegraphics[width=1.05\linewidth]{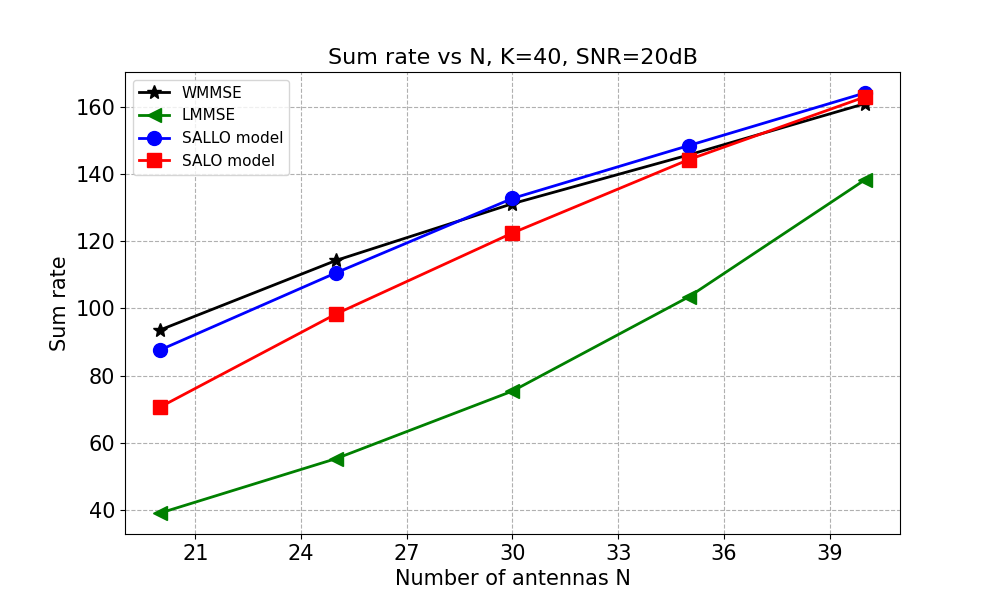}
    \caption{Performance comparison between a SALLO model and a SALO model.}
    \label{comp_lift_model}
\end{figure}

We then compare the proposed \ac{SALLO} model with the \ac{SALO} model described in Remark~\ref{remark_1}, to justify the effect of \textbf{representation lifting} in \ac{SALLO}. Specifically, under Gaussian channels with $\text{SNR}=20$ dB and $K=40$, Fig.~\ref{comp_lift_model} illustrates the testing sum rate versus the number of activated antennas $N$ for both models. It shows that \ac{SALLO} consistently outperforms \ac{SALO}, indicating that updating the auxiliary variables at each layer, which acts as a representation lifting, enhances the model’s expressivity and yields higher-quality beamformer solutions. This performance gap becomes more pronounced in overloaded regimes. On the other hand, in underloaded systems, the performance of \ac{SALO} closely approaches that of \ac{SALLO}. Hence the \ac{SALO} model can serve as a reduced-complexity alternative in underloaded systems. Nevertheless, the \ac{SALLO} scheme remains essential for achieving high-quality beamforming solutions across a broad range of system configurations, particularly in overloaded settings.

\begin{figure}
    \centering
    \includegraphics[width=1.06\linewidth]{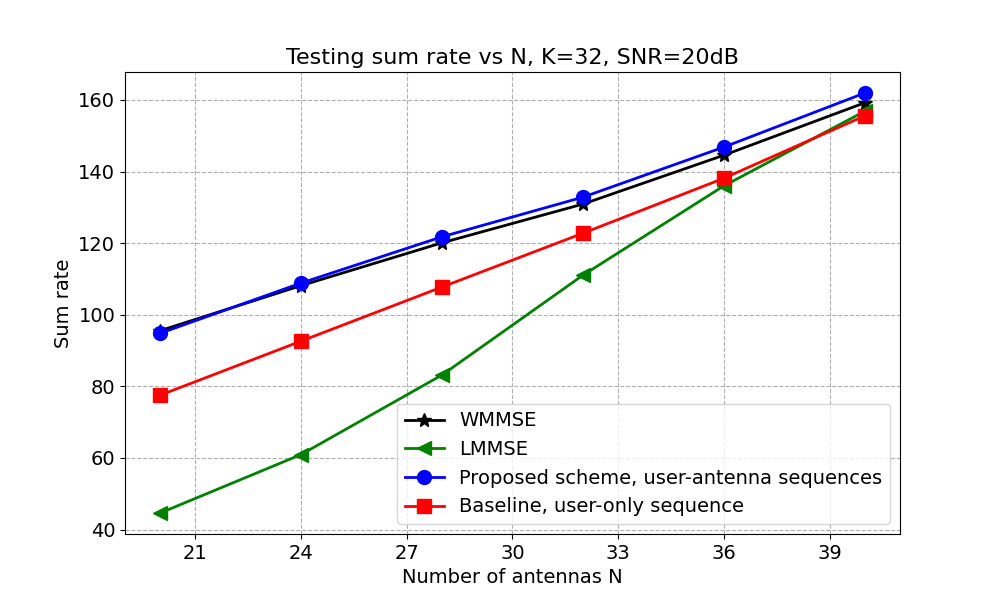}
    \caption{Performance comparison between the proposed model with user-antenna token sequences and the baseline model with user-only token sequence.}
    \label{dual_token_verify}
\end{figure}

We further validate the effectiveness of the \textbf{user-antenna dual tokenization} design in the SALLO-M Transformer model. For comparison, a baseline model is constructed that adopts only the user-level token sequence in each Transformer layer, i.e., without the antenna-level sequence $\bm{S}_{\text{a}}$ and its corresponding branch in Fig.~\ref{arch_transformer}. Both the proposed model and the baseline are trained using the same CL with random masking strategy described in Sec.~\ref{transformer_train_strategy} and evaluated under Gaussian channels with $\text{SNR}=20 \text{dB}$. Fig.~\ref{dual_token_verify} shows the testing sum-rate versus the number of activated antennas $N$ with $K=32$. The proposed scheme consistently outperforms the baseline. This improvement can be attributed to the additional antenna-level attention, which captures spatial correlations across antennas and provides richer structural information for beamforming optimization. In addition, dual tokenization naturally supports the proposed 2D masking operations, enabling generalization to varying user-antenna configurations. Notably, the advantage becomes more pronounced in overloaded regimes, where the interference structure is more complex and the learning task is more challenging.

\begin{figure} [htbp]
    \centering
    \includegraphics[width=1.07\linewidth]{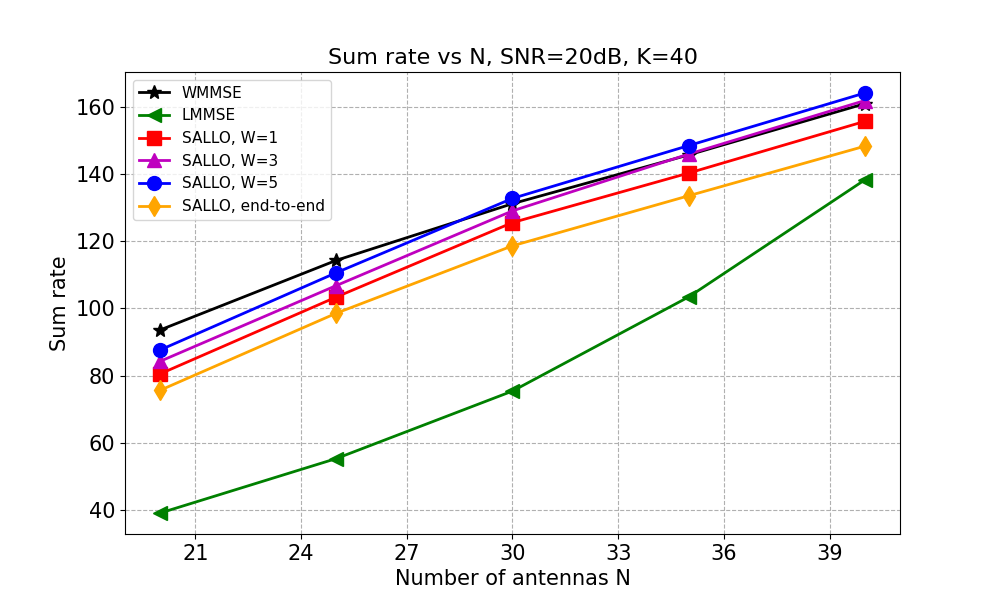}
    \caption{Performance under different sliding-window lengths.}
    \label{perf_diff_win_len}
\end{figure}

Next, we conduct an ablation study on the \textbf{sliding-window training} strategy described in Sec.~\ref{slide_window_method}. Three SALLO-M Transformer models are trained with window lengths $W=1,3,5$, respectively, and evaluated on Gaussian channel samples with $\text{SNR}=20\text{dB}$ and $K=40$, while varying the number of activated antennas $N$. In addition, a SALLO-M Transformer model trained end-to-end without a sliding window, is also included. Fig.~\ref{perf_diff_win_len} plots the testing sum rate versus the number of activated antennas $N$ for all schemes. The results show that the model with $W=5$ achieves the best performance among all variants, indicating that end-to-end training over $T$ layers is challenging and may lead to performance degradation, while a too small window size (e.g., $W=1$) fails to exploit the benefits of joint layer-wise optimization and thus limits model expressivity. Consequently, $W=5$ is adopted in the subsequent simulations.

\begin{figure} [htbp]
    \centering
    \includegraphics[width=1.06\linewidth]{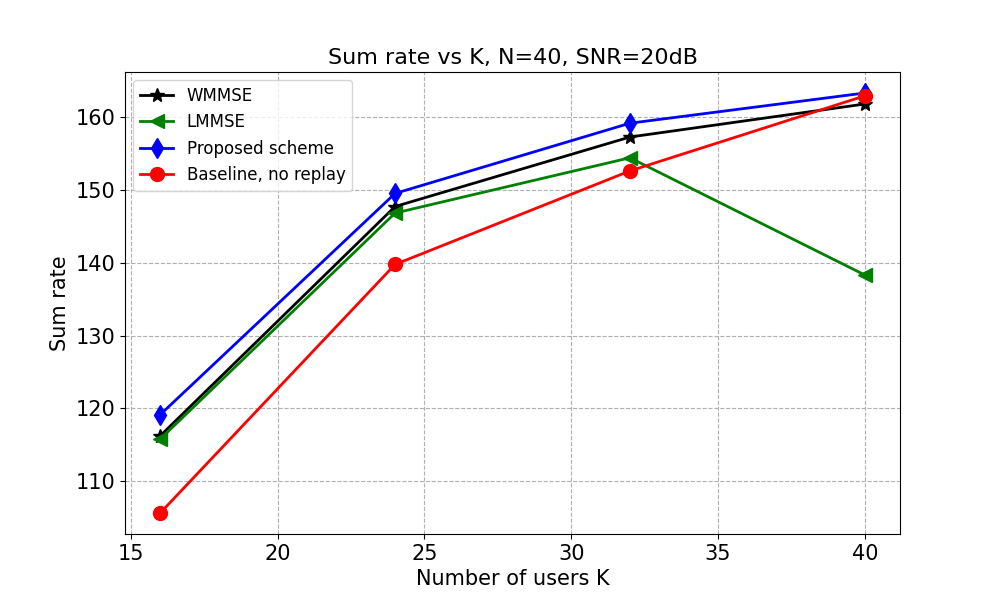}
    \caption{Performance comparison between the proposed scheme with sample replay and the baseline without sample replay.}
    \label{replay_ablation}
\end{figure}

We then proceed to examine the effectiveness of the \textbf{sample replay} strategy described in Sec.~\ref{sam_replay}. For comparison, a baseline model trained without sample replay is considered. Both models are trained and evaluated under Gaussian channels with $\text{SNR}=20 \ \text{dB}$. Fig.~\ref{replay_ablation} presents the testing sum rate versus the number of activated users $K$ with $N=40$. The proposed scheme significantly outperforms the no-replay baseline for small $K$, while the performance gap diminishes at $(K,N)=(40,40)$. This observation indicates that the baseline model suffers from forgetting earlier samples during CL-based training, whereas sample replay effectively mitigates this issue.

\begin{figure} [htbp]
    \centering
    \includegraphics[width=1.05\linewidth]{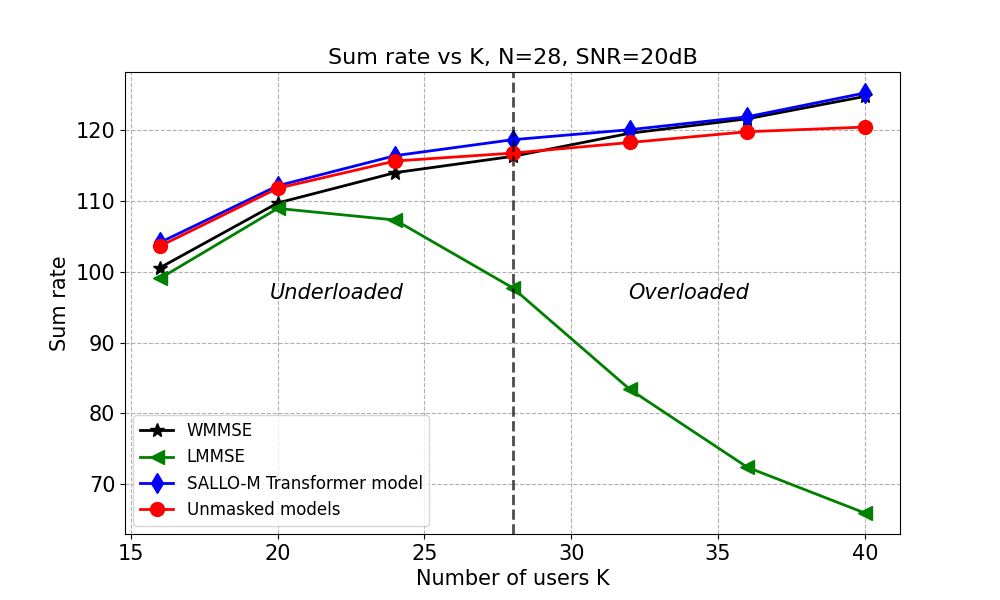}
    \caption{Performance comparison between one SALLO-M Transformer model and multiple unmasked models.}
    \label{comp_mt_dt_model}
\end{figure}

We further demonstrate the improvement of generalizability brought by the \textbf{CL method with random masking} in Sec.~\ref{cl_random_mask}. To this end, we introduce a baseline model termed the \emph{unmasked} model, which shares the same network architecture as the SALLO-M Transformer model (except for all masking components) but is trained and tested based on only one system configuration $(K,N)$. Specifically, under Gaussian channels with $\text{SNR}=20$ dB and $N=28$, Fig.~\ref{comp_mt_dt_model} shows the testing sum rate versus $K$ using a single SALLO-M Transformer model, and compares it with the performance of multiple unmasked models individually trained for each value of $K$. It is observed that a single SALLO-M Transformer consistently outperforms multiple unmasked models across all regimes, with the advantage becoming more pronounced in overloaded cases. This suggests that the training scheme proposed in Sec.~\ref{cl_random_mask} not only enables generalizability to varying configurations, but also serves as a sample-based curriculum learning paradigm that enhances sum-rate performance, particularly under challenging scenarios. Note that the performance gap between the WMMSE and LMMSE baselines becomes increasingly pronounced as $K$ grows, indicating that learning becomes more challenging --- the proposed scheme adopts the LMMSE beamformer as the initialization, which leads to slower convergence in overloaded systems. In addition, interference suppression becomes intrinsically more difficult in this scenario. 

\begin{figure} [htbp]
    \centering
    \includegraphics[width=1.05\linewidth]{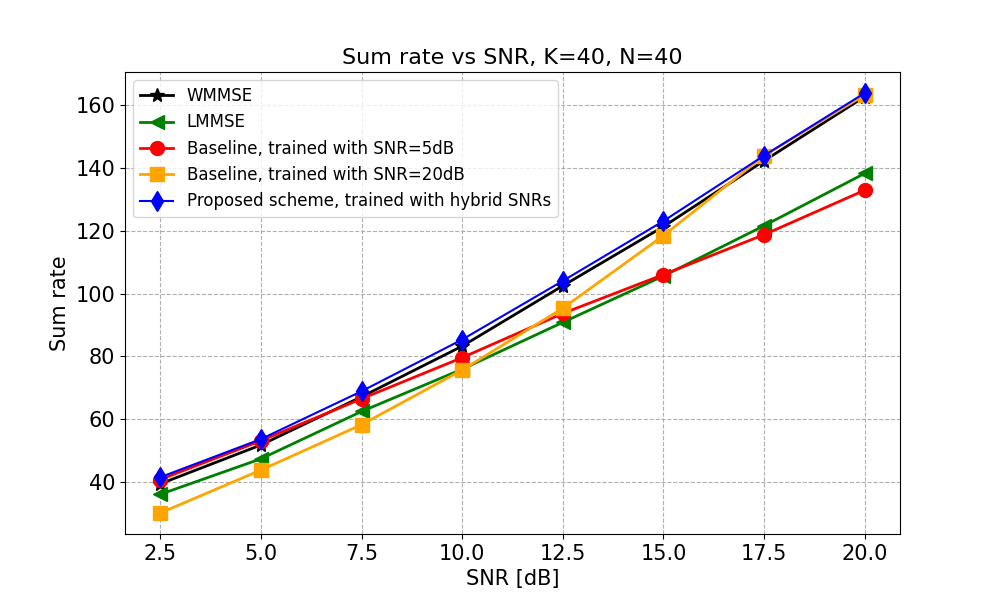}
    \caption{Performance comparison between the proposed model and multiple baselines trained with fixed SNR.}
    \label{snr_comp}
\end{figure}

Finally, we train the proposed model with a hybrid SNR set 
$\mathcal{V}_{\text{train}} = [5, 10, 15, 20] \ \text{dB}$ and then evaluate it under the SNR set $\mathcal{V}_\text{test}=[2.5,5,\dots,17.5,20] \ \text{dB}$. For comparison, two baseline models trained at fixed SNRs of $5$ dB and $20$ dB are considered. Specifically, under Gaussian channels with $K=40$ and $N=40$, Fig.~\ref{snr_comp} presents the testing sum rate versus SNR for the proposed model, along with comparisons to the two baseline models. The results show that the proposed model consistently outperforms the WMMSE benchmark, and this advantage persists on test samples corresponding to unseen SNR values during training, demonstrating strong generalizability across varying SNR conditions. In contrast, the models trained at fixed SNRs exhibit noticeable performance degradation when the testing SNR deviates from the training SNR, highlighting the importance of hybrid-SNR training.

\subsection{Overall Performance} \label{overall_result}

\begin{figure} [htbp]
    \centering
    \includegraphics[width=0.94\linewidth]{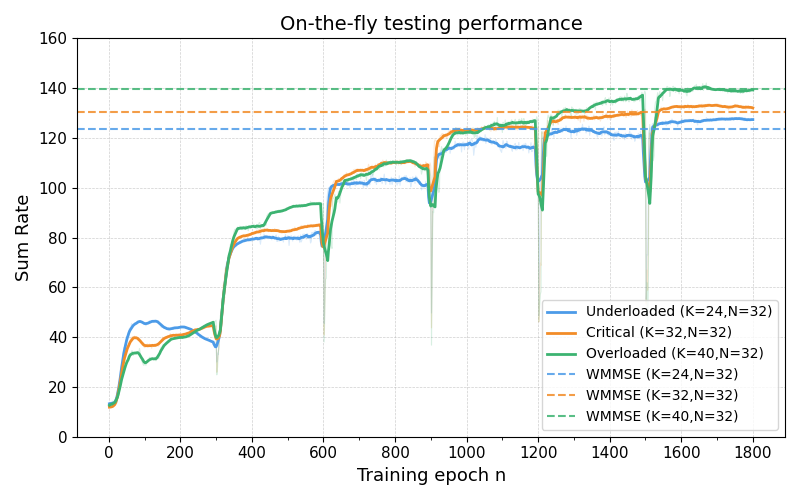}
    \caption{On-the-fly testing performance of the SALLO-M Transformer model under different channel configurations.}
    \label{otf_test_converge}
\end{figure}

The overall performance of the proposed \ac{SALLO} scheme is now evaluated. We first examine its \textbf{convergence behavior}. Since the SALLO-M Transformer model is trained under a curriculum involving multiple channel configurations, Fig.~\ref{otf_test_converge} plots the on-the-fly testing sum rate versus epochs for several representative configurations. Specifically, three groups of Gaussian channel samples with $(K,N)=(24,32),(32,32),(40,32)$ are randomly generated and kept fixed during testing, with $\text{SNR}=20$ dB. For all configurations, although the sum rate temporarily degrades at the sliding-window switching points, it gradually improves and eventually approaches (in overloaded systems) or surpasses (in underloaded systems) the WMMSE performance. In addition, the overloaded configuration approaches the WMMSE performance more slowly, indicating the increased learning difficulty in this scenario.

\begin{figure} [htbp]
    \centering
    \includegraphics[width=1.07\linewidth]{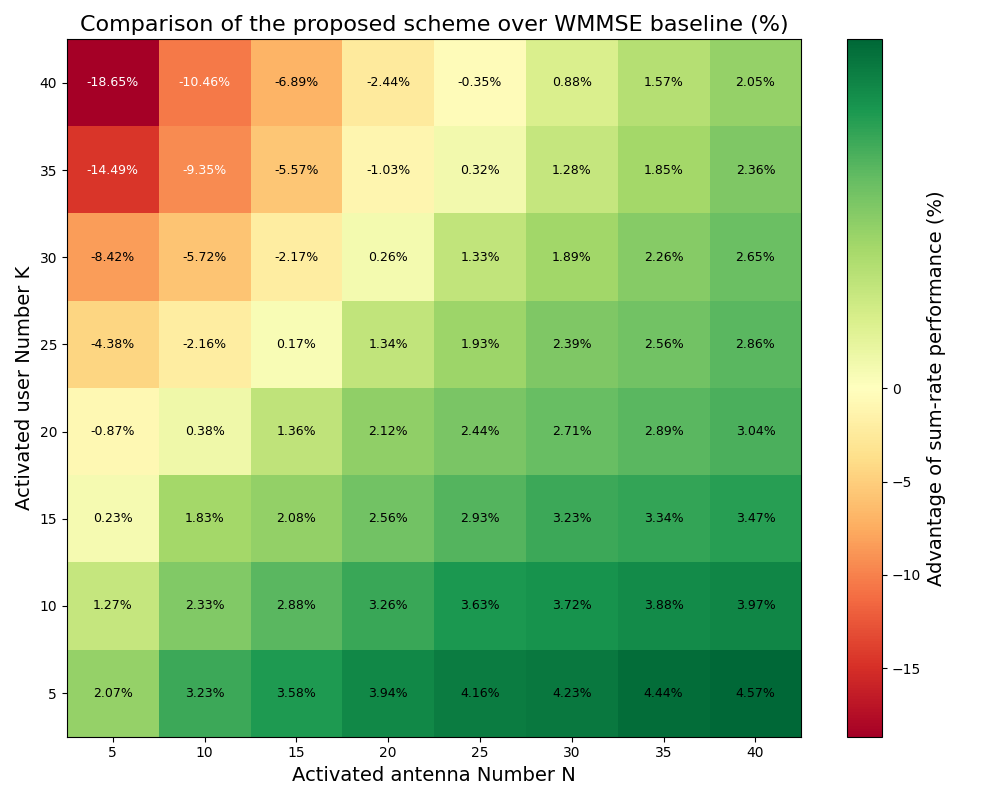}
    \caption{Sum-rate performance gain of the trained SALLO-M Transformer model relative to the WMMSE baseline under various user-antenna configurations.}
    \label{2D_advantage_heatmap}
\end{figure}

Next, under Gaussian channels with $\text{SNR}=20$ dB, and the upper bound of the sizes $\bar{K}=\bar{N}=40$, Fig.~\ref{2D_advantage_heatmap} presents a 2D heatmap of the testing performance gain of the proposed scheme over the WMMSE benchmark. To strike a balance between comprehensive evaluation and visualization clarity, we test the proposed model under all configurations $(K,N) \in [\bar{K}] \times [\bar{N}]$, while presenting the 2D results with a granularity of 5 users/antennas. It can be observed that the SALLO-M Transformer model maintains strong performance on test samples corresponding to \textbf{unseen system configurations} $(K,N)\in[\bar{K}]\times[\bar{N}]$ during training, demonstrating its generalizability within a fixed size upper bound. Notably, the proposed model outperforms the WMMSE baseline in underloaded systems, i.e., when $\frac{K}{N} \leq 1$, and even in overloaded systems when the overloading factor is below a threshold, e.g., $\frac{K}{N} \leq 2$ for $N=10$, or $\frac{K}{N} \leq 1.5$ for $N=20$, thereby demonstrating its superiority over a sufficiently broad range of practical system configurations.

We further compare the proposed model with the following two learning-assisted baselines,
\begin{itemize}
    \item \textbf{IAIDNN}: An unfolded WMMSE learning scheme \cite{hu2020iterative};
    \item \textbf{RNN Optimizer}: An online gradient-based recurrent optimization scheme \cite{johnston2024rnn};
\end{itemize}
and the following three learning-based baselines:
\begin{itemize}
    \item \textbf{Graph Transformer}: A Transformer model that adopts GNN-guided designs, exhibiting an improved size-generalizability \cite{duan2025learning}; 
    \item \textbf{CNN-Transformer}: A beamforming scheme where CNN first captures the local features and Transformer then learns the global dependencies \cite{zhang2024transformer};
    \item \textbf{LLM-adapter}: A precoding framework based on a pretrained GPT-2 model with task-specific adapters designed for various system scenarios \cite{xu2025llm}.
\end{itemize}

\begin{figure*}[htbp]
\centering
\begin{minipage}[t]{0.49\textwidth}
    \centering
     \includegraphics[width=1.08\linewidth]{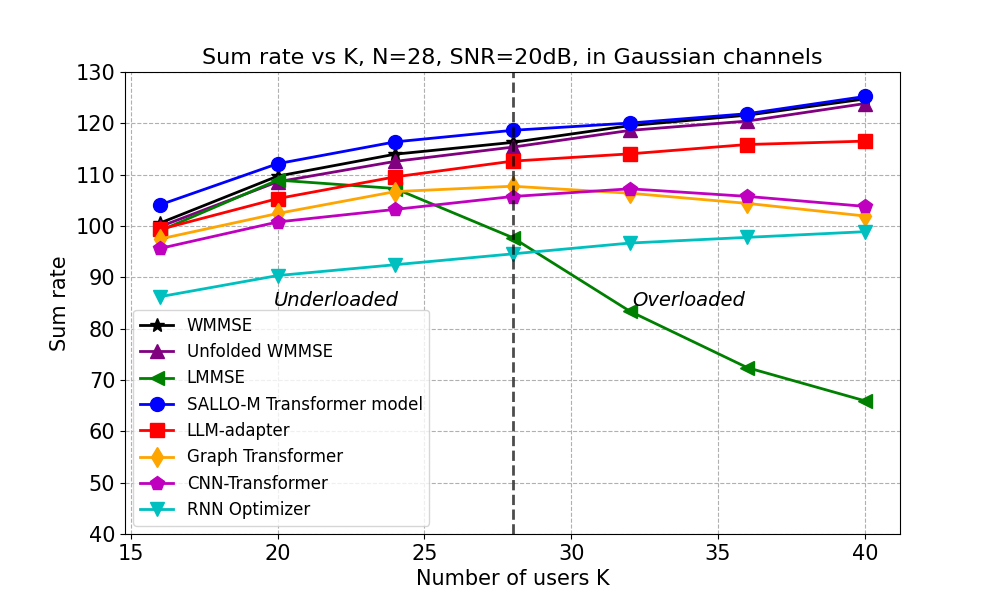}
    \caption{Comparison between the proposed scheme and baselines under varying user numbers in Gaussian channels.}
    \label{baseline_comp_gauss}
\end{minipage}
\hfill
\begin{minipage}[t]{0.49\textwidth}
    \centering
    \includegraphics[width=1.08\linewidth]{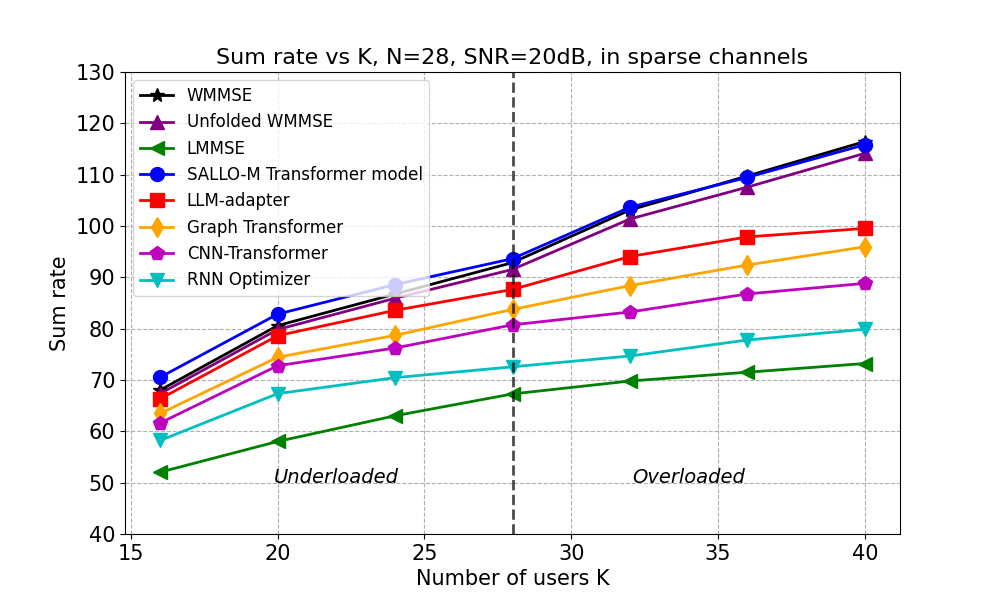}
    \caption{Comparison between the proposed scheme and baselines under varying user numbers in sparse channels.}
    \label{baseline_comp_sparse}
\end{minipage}
\end{figure*}

Fig.~\ref{baseline_comp_gauss} and Fig.~\ref{baseline_comp_sparse} illustrate the testing sum rate versus the number of served users $K$ for different beamforming schemes, with $N=28$ and $\text{SNR}=20$ dB, \textbf{under both Gaussian channel and single-cell sparse channel models,} respectively. To ensure a fair comparison, all learning-based methods are trained under the curriculum described in Sec.~\ref{cl_random_mask}, where both $K$ and $N$ are progressively increased and $(K,N) \in \mathcal{K} \times \mathcal{N}$. In contrast, the two learning-assisted methods\footnote{Learning-based methods directly learn a mapping from inputs to outputs via data-driven DL models, often replacing conventional optimization procedures. In contrast, learning-assisted methods leverage DL to guide or accelerate optimization algorithms, rather than replacing them entirely.} lack generalizability to varying problem sizes and are therefore trained separately for each test configuration $(K,N)$. As observed in Fig.~\ref{baseline_comp_gauss} and Fig.~\ref{baseline_comp_sparse}, in the underloaded regime, the proposed SALLO-M Transformer scheme consistently outperforms all baseline methods, including the WMMSE algorithm, across all system configurations and channel models. In the overloaded regime, the proposed scheme approaches the WMMSE baseline, while exhibiting a more pronounced advantage over other deep learning baselines, highlighting its superior robustness under challenging scenarios. Notably, beamforming optimization becomes significantly more difficult in overloaded sparse channels, where an excessive number of users tend to exhibit highly overlapping angular supports, making interference suppression markedly harder. To address this issue, we initialize the beamformer using the result of a two-step WMMSE iteration in this scenario, thereby alleviating the learning difficulty at the cost of a slight reduction in inference speed.

Finally, Table~\ref{inf_time} reports the GPU \textbf{inference time} (RTX 5060, 16GB memory) of all schemes shown in Fig.~\ref{baseline_comp_gauss}, evaluated with a batch size of $128$, upper bounds $(\bar{K},\bar{N})=(40,40)$, and a configuration $(K,N)=(40,28)$, under both Gaussian and sparse channels. The results show that the proposed scheme achieves faster inference than the online optimization schemes, namely ``WMMSE'', ``IAIDNN'', and ``RNN-optimizer'' baselines, since the proposed scheme shifts most online computations to the offline training. It also outperforms the ``LLM-adapter'' baseline in terms of inference overhead, owing to a more lightweight model. In contrast, the proposed scheme is slightly slower than other Transformer-based methods, namely ``Graph Transformer'' and ``CNN-Transformer'' baselines, due to the additional gradient ascent steps during inference. Notably, the inference speed of the proposed scheme is significantly slower in overloaded sparse channels because of the use of higher-quality beamformer initialization, which incurs additional online computational overhead.

\begin{table*}[htbp]
\centering
\caption{GPU inference time (seconds) per batch with $(K,N)=(40,28)$ and $\mathrm{SNR}=20 \ \mathrm{dB}$.}
\label{inf_time}
\begin{tabular}{c|ccccccc}
\hline
\textbf{Channel} & \textbf{WMMSE} & \textbf{RNN Optimizer} & \textbf{IAIDNN} & \textbf{LLM-adapter} & \textbf{SALLO-M} & \textbf{Graph Transformer} & \textbf{CNN-Transformer} \\
\hline
Gaussian & 8.782 & 4.753 & 1.327 & 0.148 & 0.032 & 0.021 & 0.012 \\
Sparse   & 10.326 & 5.123 & 1.542 & 0.206 & 0.124 & 0.032 & 0.024 \\
\hline
\end{tabular}
\end{table*}

\section{conclusions} \label{sec_conclu} 

This paper proposes a semi-amortized lifted learning-to-optimize masked (SALLO-M) Transformer model for scalable and generalizable downlink beamforming in MU-MISO systems. By jointly updating auxiliary variables and beamformer solutions across layers and refining them with a few gradient steps, the proposed architecture achieves scalable performance with fast inference. Dual tokenization and masked Transformer designs enable a single model to generalize across varying user and antenna configurations without retraining. In addition, sliding-window training, curriculum learning with random masking, and sample replay strategies improve training stability and convergence. Simulation results under both Gaussian and sparse channel models show that the proposed scheme consistently outperforms existing deep learning baselines across diverse settings, including challenging overloaded regimes, while being significantly faster than state-of-the-art online iterative optimization algorithms. Future work will explore broader channel models, as well as robustness to imperfect channel information and model mismatch.

\bibliographystyle{IEEEtran}
\bibliography{IEEEabrv,bibfile}

@article{bjornson2014optimal,
  title={Optimal multiuser transmit beamforming: A difficult problem with a simple solution structure [lecture notes]},
  author={Bj{\"o}rnson, Emil and Bengtsson, Mats and Ottersten, Bj{\"o}rn},
  journal={IEEE Signal Processing Magazine},
  volume={31},
  number={4},
  pages={142--148},
  year={2014},
  publisher={IEEE}
}

@article{johnston2023curriculum,
  title={A Curriculum Learning Approach to Optimization with Application to Downlink Beamforming},
  author={Johnston, Jeremy and Liu, Xiao-Yang and Wu, Shixun and Wang, Xiaodong},
  journal={IEEE Transactions on Signal Processing},
  year={2023},
  publisher={IEEE}
}

@article{johnston2024rnn,
  title={{RNN} Beamforming Optimizer for Rate-Splitting Multiple Access and Cell-Free Massive {MIMO}},
  author={Johnston, Jeremy and Wang, Xiaodong},
  journal={IEEE Transactions on Communications},
  year={2024},
  publisher={IEEE}
}

@article{lozano2006optimum,
  title={Optimum power allocation for parallel {G}aussian channels with arbitrary input distributions},
  author={Lozano, Angel and Tulino, Antonia M and Verd{\'u}, Sergio},
  journal={IEEE Transactions on Information Theory},
  volume={52},
  number={7},
  pages={3033--3051},
  year={2006},
  publisher={IEEE}
}

@article{shi2011iteratively,
  title={An iteratively weighted MMSE approach to distributed sum-utility maximization for a {MIMO} interfering broadcast channel},
  author={Shi, Qingjiang and Razaviyayn, Meisam and Luo, Zhi-Quan and He, Chen},
  journal={IEEE Transactions on Signal Processing},
  volume={59},
  number={9},
  pages={4331--4340},
  year={2011},
  publisher={IEEE}
}

@article{bjornson2013optimal,
  title={Optimal resource allocation in coordinated multi-cell systems},
  author={Bj{\"o}rnson, Emil and Jorswieck, Eduard and others},
  journal={Foundations and Trends{\textregistered} in Communications and Information Theory},
  volume={9},
  number={2--3},
  pages={113--381},
  year={2013},
  publisher={Now Publishers, Inc.}
}

@article{li2017learning,
  title={Learning to optimize neural nets},
  author={Li, Ke and Malik, Jitendra},
  journal={arXiv preprint arXiv:1703.00441},
  year={2017}
}

@inproceedings{wichrowska2017learned,
  title={Learned optimizers that scale and generalize},
  author={Wichrowska, Olga and Maheswaranathan, Niru and Hoffman, Matthew W and Colmenarejo, Sergio Gomez and Denil, Misha and Freitas, Nando and Sohl-Dickstein, Jascha},
  booktitle={International Conference on Machine Learning},
  pages={3751--3760},
  year={2017},
  organization={PMLR}
}

@article{zhang2024transformer,
  title={{T}ransformer-based predictive beamforming for integrated sensing and communication in vehicular networks},
  author={Zhang, Yunwu and Li, Shibao and Li, Dongyang and Zhu, Jinze and Guan, Qishuai},
  journal={IEEE Internet of Things Journal},
  volume={11},
  number={11},
  pages={20690--20705},
  year={2024},
  publisher={IEEE}
}

@article{li2024hpe,
  title={{HPE} transformer: Learning to optimize multi-group multicast beamforming under nonconvex {Q}o{S} constraints},
  author={Li, Yang and Liu, Ya-Feng},
  journal={IEEE Transactions on Communications},
  volume={72},
  number={9},
  pages={5581--5594},
  year={2024},
  publisher={IEEE}
}

@article{ting2024adaptive,
  title={Adaptive {TTD} Configurations for Near-Field Communications: An Unsupervised {T}ransformer Approach},
  author={Ting, Hsienchih and Wang, Zhaolin and Liu, Yuanwei},
  journal={IEEE Transactions on Wireless Communications},
  year={2024},
  publisher={IEEE}
}

@inproceedings{he2016deep,
  title={Deep residual learning for image recognition},
  author={He, Kaiming and Zhang, Xiangyu and Ren, Shaoqing and Sun, Jian},
  booktitle={Proceedings of the IEEE Conference on Computer Vision and Pattern Recognition},
  pages={770--778},
  year={2016}
}

@article{liu2022learning,
  title={Learning-based predictive beamforming for integrated sensing and communication in vehicular networks},
  author={Liu, Chang and Yuan, Weijie and Li, Shuangyang and Liu, Xuemeng and Li, Husheng and Ng, Derrick Wing Kwan and Li, Yonghui},
  journal={IEEE Journal on Selected Areas in Communications},
  volume={40},
  number={8},
  pages={2317--2334},
  year={2022},
  publisher={IEEE}
}

@inproceedings{bengio2009curriculum,
  title={Curriculum learning},
  author={Bengio, Yoshua and Louradour, J{\'e}r{\^o}me and Collobert, Ronan and Weston, Jason},
  booktitle={Proceedings of the 26th annual International Conference on Machine Learning},
  pages={41--48},
  year={2009}
}

@inproceedings{cheng2022masked,
  title={Masked-attention mask {T}ransformer for universal image segmentation},
  author={Cheng, Bowen and Misra, Ishan and Schwing, Alexander G and Kirillov, Alexander and Girdhar, Rohit},
  booktitle={Proceedings of the IEEE/CVF Conference on Computer Vision and Pattern Recognition},
  pages={1290--1299},
  year={2022}
}

@inproceedings{kim2018semi,
  title={Semi-amortized variational autoencoders},
  author={Kim, Yoon and Wiseman, Sam and Miller, Andrew and Sontag, David and Rush, Alexander},
  booktitle={International Conference on Machine Learning},
  pages={2678--2687},
  year={2018},
  organization={PMLR}
}

@misc{zhang2025encoderdecodernetworkbeamformingsparse,
  title={An Encoder-Decoder Network for Beamforming over Sparse Large-Scale {MIMO} Channels}, 
  author={Yubo Zhang and Jeremy Johnston and Xiaodong Wang},
  year={2025},
  eprint={2510.02355},
  archivePrefix={arXiv},
  primaryClass={eess.SY},
  url={https://arxiv.org/abs/2510.02355}, 
}

@article{toneva2018empirical,
  title={An empirical study of example forgetting during deep neural network learning},
  author={Toneva, Mariya and Sordoni, Alessandro and Combes, Remi Tachet des and Trischler, Adam and Bengio, Yoshua and Gordon, Geoffrey J},
  journal={arXiv preprint arXiv:1812.05159},
  year={2018}
}

@article{rajapaksha2023minimizing,
  title={Minimizing energy consumption in {MU}-{MIMO} via antenna muting by neural networks with asymmetric loss},
  author={Rajapaksha, Nuwanthika and Mohammadi, Jafar and Wesemann, Stefan and Wild, Thorsten and Rajatheva, Nandana},
  journal={IEEE Transactions on Vehicular Technology},
  volume={73},
  number={5},
  pages={6600--6613},
  year={2023},
  publisher={IEEE}
}

@article{joudeh2017rate,
  title={Rate-splitting for max-min fair multigroup multicast beamforming in overloaded systems},
  author={Joudeh, Hamdi and Clerckx, Bruno},
  journal={IEEE Transactions on Wireless Communications},
  volume={16},
  number={11},
  pages={7276--7289},
  year={2017},
  publisher={IEEE}
}

@article{zhou2023algorithms,
  title={What algorithms can {T}ransformers learn? a study in length generalization},
  author={Zhou, Hattie and Bradley, Arwen and Littwin, Etai and Razin, Noam and Saremi, Omid and Susskind, Josh and Bengio, Samy and Nakkiran, Preetum},
  journal={arXiv preprint arXiv:2310.16028},
  year={2023}
}

@article{xu2025llm,
  title={{LLM}-empowered near-field communications for low-altitude economy},
  author={Xu, Zhuo and Zheng, Tianyue and Dai, Linglong},
  journal={IEEE Transactions on Communications},
  year={2025},
  publisher={IEEE}
}

@article{sohrabi2021deep,
  title={Deep learning for distributed channel feedback and multiuser precoding in {FDD} massive {MIMO}},
  author={Sohrabi, Foad and Attiah, Kareem M and Yu, Wei},
  journal={IEEE Transactions on Wireless Communications},
  volume={20},
  number={7},
  pages={4044--4057},
  year={2021},
  publisher={IEEE}
}

@article{li2025bert4beam,
  title={{BERT}4beam: Large {AI} model enabled generalized beamforming optimization},
  author={Li, Yuhang and Lu, Yang and Chen, Wei and Ai, Bo and Ding, Zhiguo and Niyato, Dusit},
  journal={arXiv preprint arXiv:2509.11056},
  year={2025}
}

@article{wen2026wifo,
  title={{W}i{F}o-{E}: A {S}calable {W}ireless {F}oundation {M}odel for {E}nd-to-{E}nd {FDD} {P}recoding in {C}ommunication {N}etworks},
  author={Wen, Weibo and Gao, Shijian and Zhang, Haotian and Cheng, Xiang and Yang, Liuqing},
  journal={arXiv preprint arXiv:2601.09186},
  year={2026}
}

@article{elbir2019hybrid,
  title={Hybrid precoding for multiuser millimeter wave massive {MIMO} systems: A deep learning approach},
  author={Elbir, Ahmet M and Papazafeiropoulos, Anastasios K},
  journal={IEEE Transactions on Vehicular Technology},
  volume={69},
  number={1},
  pages={552--563},
  year={2019},
  publisher={IEEE}
}

@article{rolnick2019experience,
  title={Experience replay for continual learning},
  author={Rolnick, David and Ahuja, Arun and Schwarz, Jonathan and Lillicrap, Timothy and Wayne, Gregory},
  journal={Advances in Neural Information Processing Systems},
  volume={32},
  year={2019}
}

@inproceedings{he2022masked,
  title={Masked autoencoders are scalable vision learners},
  author={He, Kaiming and Chen, Xinlei and Xie, Saining and Li, Yanghao and Doll{\'a}r, Piotr and Girshick, Ross},
  booktitle={Proceedings of the IEEE/CVF Conference on Computer Vision and Pattern Recognition},
  pages={16000--16009},
  year={2022}
}

@inproceedings{devlin2019bert,
  title={{BERT}: Pre-training of Deep Bidirectional {T}ransformers for Language Understanding},
  author={Devlin, Jacob and Chang, Ming-Wei and Lee, Kenton and Toutanova, Kristina},
  booktitle={Proceedings of NAACL-HLT},
  pages={4171--4186},
  year={2019}
}

@article{vaswani2017attention,
  title={Attention is all you need},
  author={Vaswani, Ashish and Shazeer, Noam and Parmar, Niki and Uszkoreit, Jakob and Jones, Llion and Gomez, Aidan N and Kaiser, {\L}ukasz and Polosukhin, Illia},
  journal={Advances in Neural Information Processing Systems},
  volume={30},
  year={2017}
}

@inproceedings{fan2021mask,
  title={Mask Attention Networks: Rethinking and Strengthening {T}ransformer},
  author={Fan, Zhiwei and Gong, Yizhe and Liu, Dayiheng and Wei, Zihan and Wang, Siyuan and Jiao, Jianbo and Duan, Nan and Zhang, Ruofei and Huang, Xuanjing},
  booktitle={Proc. NAACL},
  pages={1692--1701},
  year={2021}
}

@article{lavi2023learn,
  title={Learn to rapidly and robustly optimize hybrid precoding},
  author={Lavi, Ortal and Shlezinger, Nir},
  journal={IEEE Transactions on Communications},
  volume={71},
  number={10},
  pages={5814--5830},
  year={2023},
  publisher={IEEE}
}

@article{li2024GNN,
  title={{GNN}-based beamforming for sum-rate maximization in {MU-MISO} networks},
  author={Li, Yuhang and Lu, Yang and Ai, Bo and Dobre, Octavia A and Ding, Zhiguo and Niyato, Dusit},
  journal={IEEE Transactions on Wireless Communications},
  volume={23},
  number={8},
  pages={9251--9264},
  year={2024},
  publisher={IEEE}
}

@article{duan2025learning,
  title={Learning Precoding in Multi-User Multi-Antenna Systems: {T}ransformer or Graph {T}ransformer?},
  author={Duan, Yuxuan and Guo, Jia and Yang, Chenyang},
  journal={IEEE Transactions on Wireless Communications},
  volume={25},
  pages={6284--6300},
  year={2025},
  publisher={IEEE}
}

@inproceedings{carreira2014distributed,
  title={Distributed optimization of deeply nested systems},
  author={Carreira-Perpinan, Miguel and Wang, Weiran},
  booktitle={Artificial Intelligence and Statistics},
  pages={10--19},
  year={2014},
  organization={PMLR}
}

@article{bengio2006greedy,
  title={Greedy layer-wise training of deep networks},
  author={Bengio, Yoshua and Lamblin, Pascal and Popovici, Dan and Larochelle, Hugo},
  journal={Advances in Neural Information Processing Systems},
  volume={19},
  year={2006}
}

@article{kim2022bipartite,
  title={A bipartite graph neural network approach for scalable beamforming optimization},
  author={Kim, Junbeom and Lee, Hoon and Hong, Seung-Eun and Park, Seok-Hwan},
  journal={IEEE Transactions on Wireless Communications},
  volume={22},
  number={1},
  pages={333--347},
  year={2022},
  publisher={IEEE}
}

@article{hu2020iterative,
  title={Iterative algorithm induced deep-unfolding neural networks: Precoding design for multiuser {MIMO} systems},
  author={Hu, Qiyu and Cai, Yunlong and Shi, Qingjiang and Xu, Kaidi and Yu, Guanding and Ding, Zhi},
  journal={IEEE Transactions on Wireless Communications},
  volume={20},
  number={2},
  pages={1394--1410},
  year={2020},
  publisher={IEEE}
}

@article{geng2021deep,
  title={Deep shearlet residual learning network for single image super-resolution},
  author={Geng, Tianyu and Liu, Xiao-Yang and Wang, Xiaodong and Sun, Guiling},
  journal={IEEE Transactions on Image Processing},
  volume={30},
  pages={4129--4142},
  year={2021},
  publisher={IEEE}
}

\end{document}